\documentclass[10pt,twocolumn,letterpaper]{article}

\usepackage{cvpr}
\usepackage{times}
\usepackage{epsfig}
\usepackage{graphicx}
\usepackage{amsmath}
\usepackage{amssymb}

\usepackage[breaklinks=true,bookmarks=false]{hyperref}

\cvprfinalcopy 

\usepackage{mathtools}

\DeclarePairedDelimiter\floor{\lfloor}{\rfloor}
\usepackage{amssymb}
\usepackage{multirow}
\usepackage{siunitx}
\newcommand{\tabincell}[2]{\begin{tabular}{@{}#1@{}}#2\end{tabular}}

\begin{document}

\title{PointRCNN: 3D Object Proposal Generation and Detection from Point Cloud}

\author{Shaoshuai Shi  ~~\quad Xiaogang Wang  ~~\quad Hongsheng Li\\
The Chinese University of Hong Kong\\
{\tt\small {\{ssshi, xgwang, hsli\}}@ee.cuhk.edu.hk}
}

\maketitle

\begin{abstract}
In this paper, we propose PointRCNN for 3D object detection from raw point cloud. The whole framework is composed of two stages: stage-1 for the bottom-up 3D proposal generation and stage-2 for refining proposals in the canonical coordinates to obtain the final detection results. Instead of generating proposals from RGB image or projecting point cloud to bird's view or voxels as previous methods do, our stage-1 sub-network directly generates a small number of high-quality 3D proposals from point cloud in a bottom-up manner via segmenting the point cloud of the whole scene into foreground points and background. The stage-2 sub-network transforms the pooled points of each proposal to canonical coordinates to learn better local spatial features, which is combined with global semantic features of each point learned in stage-1 for accurate box refinement and confidence prediction. Extensive experiments on the 3D detection benchmark of KITTI dataset show that our proposed architecture outperforms state-of-the-art methods with remarkable margins by using only point cloud as input. 
The code is available at \href{https://github.com/sshaoshuai/PointRCNN}{https://github.com/sshaoshuai/PointRCNN}.
   
\end{abstract}

\section{Introduction}

Deep learning has achieved remarkable progress on 2D computer vision tasks, including object detection \cite{girshick2015fast, ren2015faster, li2019_internet} and instance segmentation \cite{dai2016instance, he2017mask, liu2017sgn}, etc. Beyond 2D scene understanding, 3D object detection is crucial and indispensable for many real-world applications, such as autonomous driving and domestic robots. While recent developed 2D detection algorithms are capable of handling large variations of viewpoints and background clutters in images, the detection of 3D objects with point clouds still faces great challenges from the irregular data format and large search space of 6 Degrees-of-Freedom (DoF) of 3D object. 

\begin{figure}
	\begin{center}
		\includegraphics[width=1.0\linewidth]{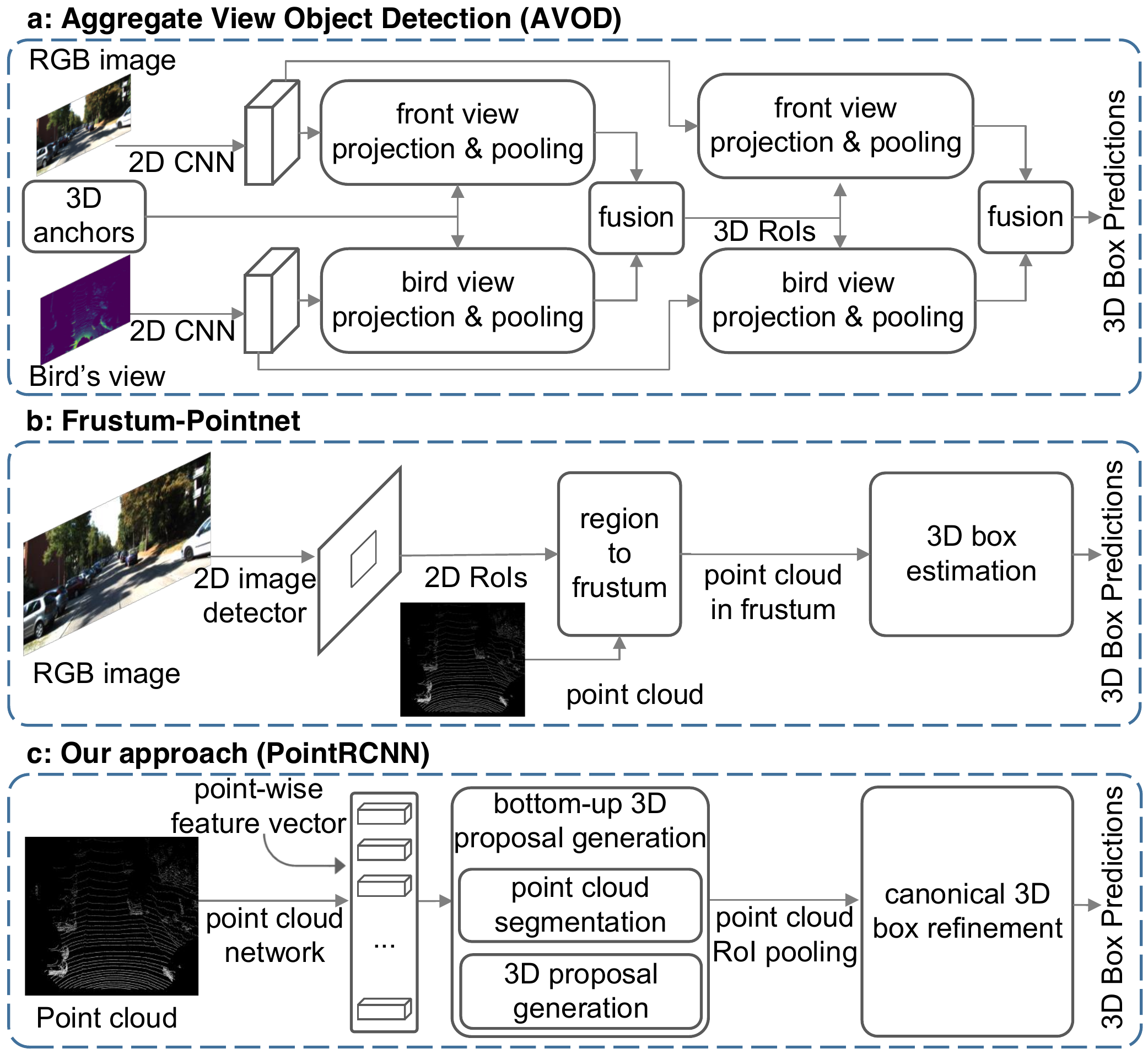}
	\end{center}
	\caption{Comparison with state-of-the-art methods. Instead of generating proposals from fused feature maps of bird's view and front view \cite{ku2017joint}, or RGB images \cite{qi2017frustum}, our method directly generates 3D proposals from raw point cloud in a bottom-up manner.}
	\label{fig:method_compare}
	\vspace{-4mm}
\end{figure}

In autonomous driving, the most commonly used 3D sensors are the LiDAR sensors, which generate 3D point clouds to capture the 3D structures of the scenes. The difficulty of point cloud-based 3D object detection mainly lies in irregularity of the point clouds. State-of-the-art 3D detection methods either leverage the mature 2D detection frameworks by projecting the point clouds into bird's view \cite{ku2017joint, yang2018pixor, liang2018deep} (see Fig.~\ref{fig:method_compare}~(a)), to the frontal view \cite{chen2017multi, xu2018multi}, or to the regular 3D voxels \cite{song2016deep, zhou2017voxelnet}, which are not optimal and suffer from information loss during the quantization.

Instead of transforming point cloud to voxels or other regular data structures for feature learning, Qi \etal \cite{qi2017pointnet, qi2017pointnet++} proposed PointNet for learning 3D representations directly from point cloud data for point cloud classification and segmentation. As shown in Fig.~\ref{fig:method_compare}~(b), their follow-up work \cite{qi2017frustum} applied PointNet in 3D object detection to estimate the 3D bounding boxes based on the cropped frustum point cloud from the 2D RGB detection results. However, the performance of the method heavily relies on the 2D detection performance and cannot take the advantages of 3D information for generating robust bounding box proposals.

Unlike object detection from 2D images, 3D objects in autonomous driving scenes are naturally and well separated by annotated 3D bounding boxes. In other words, the training data for 3D object detection directly provides the semantic masks for 3D object segmentation. This is a key difference between 3D detection and 2D detection training data. In 2D object detection, the bounding boxes could only provide weak supervisions for semantic segmentation \cite{dai2015boxsup}.

Based on this observation, we present a novel two-stage 3D object detection framework, named PointRCNN, which directly operates on 3D point clouds and achieves robust and accurate 3D detection performance (see Fig.~\ref{fig:method_compare}~(c)). The proposed framework consists of two stages, the first stage aims at generating 3D bounding box proposal in a bottom-up scheme. By utilizing 3D bounding boxes to generate ground-truth segmentation mask, the first stage segments foreground points and generates a small number of bounding box proposals from the segmented points simultaneously. Such a strategy avoids using the large number of 3D anchor boxes in the whole 3D space as previous methods \cite{zhou2017voxelnet, ku2017joint, chen2017multi} do and saves much computation. 

The second stage of PointRCNN conducts canonical 3D box refinement. After the 3D proposals are generated, a point cloud region pooling operation is adopted to pool learned point representations from stage-1.
Unlike existing 3D methods that directly estimate the global box coordinates, the pooled 3D points are transformed to the canonical coordinates and combined with the pooled point features as well as the segmentation mask from stage-1 for learning relative coordinate refinement. 
This strategy fully utilizes all information provided by our robust stage-1 segmentation and proposal sub-network. To learn more effective coordinate refinements, we also propose the full bin-based 3D box regression loss for proposal generation and refinement, and the ablation experiments show that it converges faster and achieves higher recall than other 3D box regression loss.

Our contributions could be summarized into three-fold. (1) We propose a novel bottom-up point cloud-based 3D bounding box proposal generation algorithm, which generates a small number of high-quality 3D proposals via segmenting the point cloud into foreground objects and background. The learned point representation from segmentation is not only good at proposal generation but is also helpful for the later box refinement.
(2) The proposed canonical 3D bounding box refinement takes advantages of our high-recall box proposals generated from stage-1 and learns to predict box coordinates refinements in the canonical coordinates with robust bin-based losses. 
(3) Our proposed 3D detection framework PointRCNN outperforms state-of-the-art  methods with remarkable margins and ranks first among all published works as of Nov.~16 2018 on the 3D detection test board of KITTI by using only point clouds as input.

\section{Related Work}

\noindent
\textbf{3D object detection from 2D images.}
There are existing works on estimating the 3D bounding box from images. \cite{mousavian20173d,li2019gs3d} leveraged the geometry constraints between 3D and 2D bounding box to recover the 3D object pose. \cite{chabot2017deep, zhu2014single, mottaghi2015coarse} exploited the similarity between 3D objects and the CAD models. Chen \etal \cite{chen2016monocular, chen20153d} formulated the 3D geometric information of objects as an energy function to score the predefined 3D boxes. These works can only generate coarse 3D detection results due to the lack of depth information and can be substantially affected by appearance variations.

\smallskip
\noindent
\textbf{3D object detection from point clouds.}
State-of-the-art 3D object detection methods proposed various ways to learn discriminative features from the sparse 3D point clouds.
\cite{chen2017multi, ku2017joint, yang2018pixor, liang2018deep, yang2018hdnet} projected point cloud to bird's view and utilized 2D CNN to learn the point cloud features for 3D box generation. Song \etal \cite{song2016deep} and Zhou \etal \cite{zhou2017voxelnet} grouped the points into voxels and used 3D CNN to learn the features of voxels to generate 3D boxes. However, the bird's view projection and voxelization suffer from information loss due to the data quantization, and the 3D CNN is both memory and computation inefficient. \cite{qi2017frustum, xu2017pointfusion} utilized mature 2D detectors to generate 2D proposals from images and reduced the size of 3D points in each cropped image regions. PointNet \cite{qi2017pointnet, qi2017pointnet++} is then used to learn the point cloud features for 3D box estimation. But the 2D image-based proposal generation might fail on some challenging cases that could only be well observed from 3D space. Such failures could not be recovered by the 3D box estimation step. In contrast, our bottom-to-up 3D proposal generation method directly generates robust 3D proposals from point clouds, which is both efficient and quantization free.

\smallskip
\noindent
\textbf{Learning point cloud representations.}
Instead of representing the point cloud as voxels \cite{maturana2015voxnet, riegler2017octnet, song2017semantic} or multi-view formats \cite{qi2016volumetric, su2015multi, su20153d}, Qi \etal \cite{qi2017pointnet} presented the PointNet architecture to directly learn point features from raw point clouds, which greatly increases the speed and accuracies of point cloud classification and segmentation. The follow-up works \cite{qi2017pointnet++, huang2018recurrent} further improve the extracted feature quality by considering the local structures in point clouds. Our work extends the point-based feature extractors to 3D point cloud-based object detection, leading to a novel two-stage 3D detection framework, which directly generate 3D box proposals and detection results from raw point clouds.

\begin{figure*}
	\begin{center}
		\includegraphics[width=1.0\linewidth]{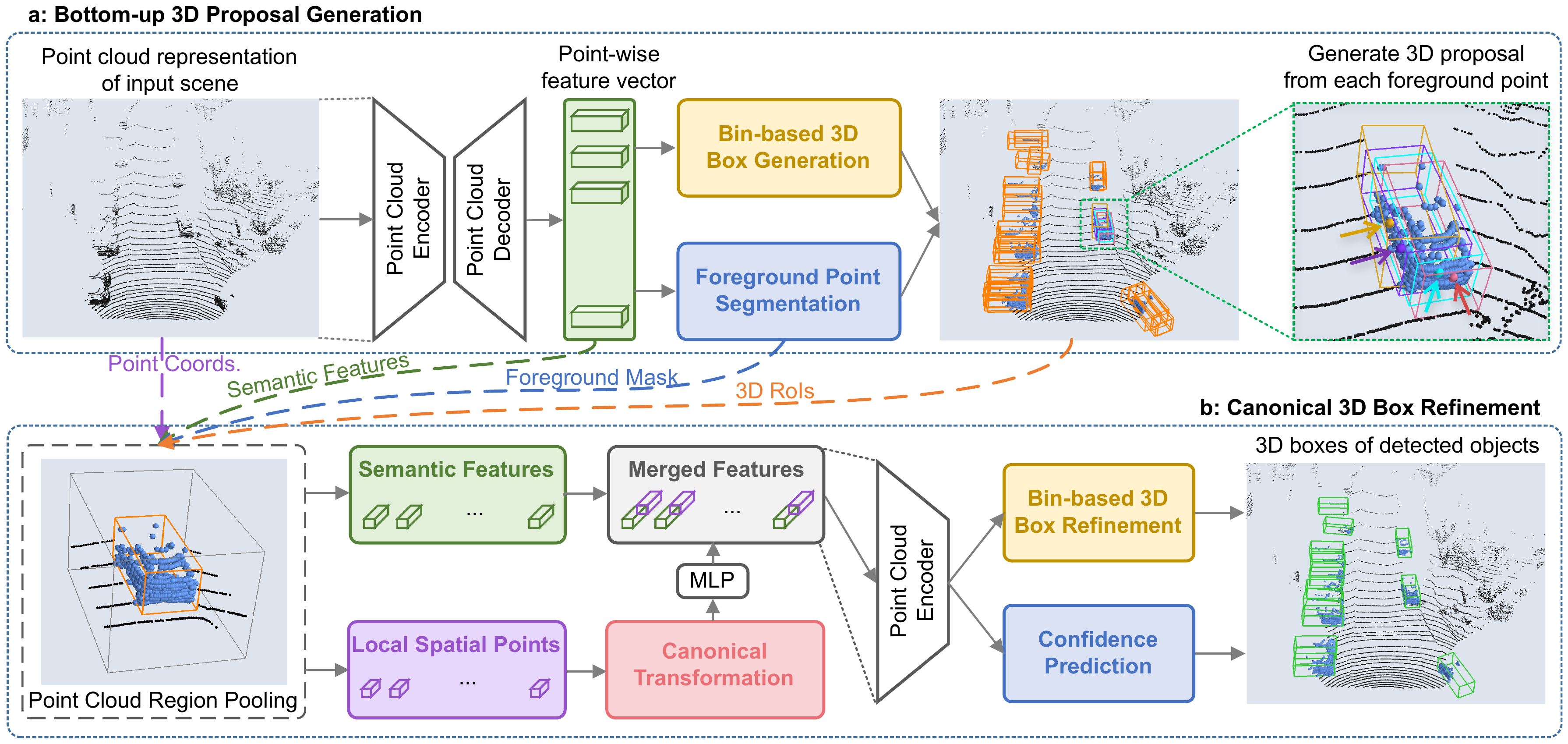}
	\end{center}
	\vspace{-2mm}
	\caption{The \textbf{PointRCNN} architecture for 3D object detection from point cloud. The whole network consists of two parts: (a) for generating 3D proposals from raw point cloud  in a bottom-up manner. (b) for refining the 3D proposals in canonical coordinate.}
	\label{fig:framework}
	\vspace{-3mm}
\end{figure*}

\section{PointRCNN for Point Cloud 3D Detection}
In this section, we present our proposed two-stage detection framework, PointRCNN, for detecting 3D objects from irregular point cloud. The overall structure is illustrated in Fig.~\ref{fig:framework},  which consists of the bottom-up 3D proposal generation stage and the canonical bounding box refinement stage. 

\subsection{Bottom-up 3D proposal generation via point cloud segmentation}\label{sec:rpn}

Existing 2D object detection methods could be classified into one-stage and two-stage methods, where one-stage methods \cite{lin2018focal, liu2016ssd, redmon2018yolov3, redmon2017yolo9000, redmon2016you} are generally faster but directly estimate object bounding boxes without refinement, while two-stage methods \cite{he2017mask, lin2017feature, ren2015faster, girshick2015fast} generate proposals firstly and further refine the proposals and confidences in a second stage. However, direct extension of the two-stage methods from 2D to 3D is non-trivial due to the huge 3D search space and the irregular format of point clouds. 
AVOD \cite{ku2017joint} places 80-100k anchor boxes in the 3D space and pool features for each anchor in multiple views for generating proposals. F-PointNet \cite{qi2017frustum} generates 2D proposals from 2D images, 
and estimate 3D boxes based on the 3D points cropped from the 2D regions, which might miss difficult objects that could only be clearly observed from 3D space.
 
We propose an accurate and robust 3D proposal generation algorithm as our stage-1 sub-network based on whole-scene point cloud segmentation.
We observe that objects in 3D scenes are naturally separated without overlapping each other. All 3D objects' segmentation masks could be directly obtained by their 3D bounding box annotations, \ie, 3D points inside 3D boxes are considered as foreground points. 

We therefore propose to generate 3D proposals in a bottom-up manner. Specifically, we learn point-wise features to segment the raw point cloud and to generate 3D proposals from the segmented foreground points simultaneously. Based on this bottom-up strategy, our method avoids using a large set of predefined 3D boxes in the 3D space and significantly constrains the search space for 3D proposal generation. The experiments show that our proposed 3D box proposal method achieves significantly higher recall than 3D anchor-based proposal generation methods.

\vspace{-0.5mm}
\smallskip
\noindent
{\bf Learning point cloud representations.} To learn discriminative point-wise features for describing the raw point clouds, we utilize the PointNet++ \cite{qi2017pointnet++} with multi-scale grouping as our backbone network.
There are several other alternative point-cloud network structures, such as \cite{qi2017pointnet, jiang2018pointsift} 
or VoxelNet~\cite{zhou2017voxelnet} with sparse convolutions~\cite{3DSemanticSegmentationWithSubmanifoldSparseConvNet}, 
which could also be adopted as our backbone network.

\vspace{-0.5mm}
\smallskip
\noindent
{\bf Foreground point segmentation.} The foreground points provide rich information on predicting their associated objects' locations and orientations. By learning to segment the foreground points, the point-cloud network is forced to capture contextual information for making accurate point-wise prediction, which is also beneficial for 3D box generation. 
We design the bottom-up 3D proposal generation method to generate 3D box proposals directly from the foreground points, \ie, the foreground segmentation and 3D box proposal generation are performed simultaneously. 

Given the point-wise features encoded by the backbone point cloud network,
we append one segmentation head for estimating the foreground mask and one box regression head for generating 3D proposals. 
For point segmentation, the ground-truth segmentation mask is naturally provided by the 3D ground-truth boxes. 
The number of foreground points is generally much smaller than that of the background points for a large-scale outdoor scene. Thus we use the focal loss \cite{lin2018focal} to handle the 
class imbalance problem as
\vspace{-2mm}
\begin{align}\label{cls_loss}
\mathcal{L}_{\textrm{focal}}(p_t)  &= -\alpha_t(1 - p_t)^\gamma \log(p_t),\\
\text{where } p_t &= 
\begin{cases}
p  &\text{for forground point}\\ 
1 - p & \text{otherwise}
\end{cases} \nonumber
\end{align}
\vspace{-4mm}

\noindent
During training point cloud segmentation, we keep the default settings $\alpha_t=0.25$ and $\gamma=2$ as the original paper.

\smallskip
\noindent
{\bf Bin-based 3D bounding box generation.} As we mentioned above, 
a box regression head is also appended for simultaneously generating bottom-up 3D proposals with the foreground point segmentation. During training, we only require the box regression head to regress 3D bounding box locations from foreground points. Note that although boxes are not regressed from the background points, those points also provide supporting information for generating boxes because of the receptive field of the point-cloud network.

A 3D bounding box is represented as $(x, y, z, h, w, l, \theta)$ in the LiDAR coordinate system, where $(x, y, z)$ is the object center location, $(h, w, l)$ is the object size, and $\theta$ is the object orientation from the bird's view. To constrain the generated 3D box proposals, we propose bin-based regression losses for estimating 3D bounding boxes of objects.

\begin{figure}[t]
	\begin{center}
		\includegraphics[width=0.99\linewidth,height=4.5cm]{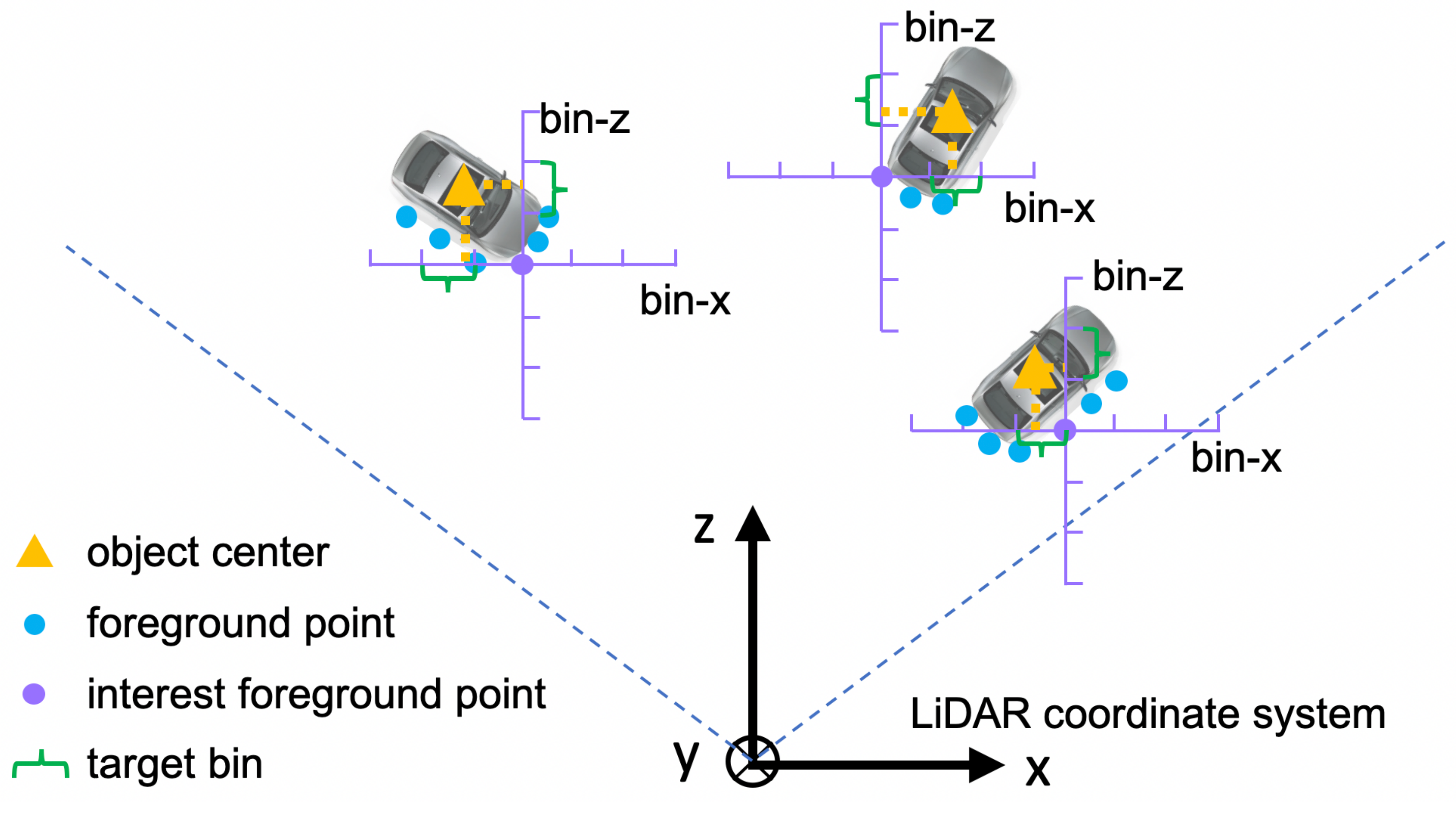}
	\end{center}
	\caption{Illustration of bin-based localization. The surrounding area along X and Z axes of each foreground point is split into a series of bins to locate the object center.}
	\label{fig:loc}
	\vspace{-5mm}
\end{figure}

For estimating center location of an object, as shown in Fig.\ \ref{fig:loc}, we split the surrounding area of each foreground point into a series of discrete bins along the $X$ and $Z$ axes.
Specifically, we set a search range $\mathcal{S}$ for each $X$ and $Z$ axis of the current foreground point, and each 1D search range is divided into bins of uniform length $\delta$ to represent different object centers $(x, z)$ on the $X$-$Z$ plane. We observe that using bin-based classification with cross-entropy loss for the $X$ and $Z$ axes instead of direct regression with smooth $L1$ loss results in more accurate and robust center localization.

The localization loss for the $X$ or $Z$ axis consists of two terms, one term for bin classification along each $X$ and $Z$ axis, and the other term for residual regression within the classified bin.
For the center location $y$ along the vertical $Y$ axis,
we directly utilize smooth $L1$ loss for the regression since most objects' $y$ values are within a very small range. Using the $L1$ loss is enough for obtaining accurate $y$ values.

The localization targets could therefore be formulated as
\vspace{-4.5mm}
\begin{align}\label{loc_loss}
\text{bin}_x^{(p)}  &= \floor*{\frac{x^p - x^{(p)} + \mathcal{S}}{\delta}},~
\text{bin}_z^{(p)}  = \floor*{\frac{z^p - z^{(p)} + \mathcal{S}}{\delta}},\nonumber
\\
\mathop{\text{res}_u^{(p)}}\limits_{u \in \{x, z\}}  &= \frac{1}{\mathcal{C}}\left( u^p - u^{(p)} + \mathcal{S} - \left(\text{bin}_u^{(p)} \cdot \delta + \frac{\delta}{2}\right)\right),
\\
\text{res}_y^{(p)} &= y^p - y^{(p)} \nonumber
\end{align}
\vspace{-5mm}

\noindent
where $(x^{(p)}, y^{(p)}, z^{(p)})$ is the coordinates of a foreground point of interest, $(x^p,y^p,z^p)$ is the center coordinates of its corresponding object ,
$\text{bin}^{(p)}_x$ and $\text{bin}^{(p)}_z$ are ground-truth bin assignments along $X$ and $Z$ axis, $\text{res}^{(p)}_x$ and $\text{res}^{(p)}_z$ are the ground-truth residual for further location refinement within the assigned bin, and $\mathcal{C}$ is the bin length for normalization.

The targets of orientation $\theta$ and size $(h,w,l)$ estimation are similar to those in \cite{qi2017frustum}. We divide the orientation $2\pi$ into $n$ bins, and calculate the bin classification target $\text{bin}_\theta^{(p)}$ and residual regression target $\text{res}_\theta^{(p)}$ in the same way 
as $x$ or $z$ prediction. 
The object size $(h, w, l)$ is directly regressed by calculating residual $(\text{res}_h^{(p)}, \text{res}_w^{(p)}, \text{res}_l^{(p)})$ w.r.t. the average object size of each class in the entire training set. 

In the inference stage, for the bin-based predicted parameters, $x$, $z$, $\theta$, we first choose the bin center with the highest predicted confidence and add the predicted residual to obtain the refined parameters. For other directly regressed parameters, including $y$, $h$, $w$, and $l$, we add the predicted residual to their initial values.

The overall 3D bounding box regression loss $\mathcal{L}_{\text{reg}}$ with different loss terms for training could then be formulated as
\vspace{-4mm}
\begin{align}
\mathcal{L}_\text{bin}^{(p)} &= \sum_{u \in \{x, z, \theta\}} ( \mathcal{F}_\text{cls}(\widehat{\text{bin}}_u^{(p)}, \text{bin}_u^{(p)})
+ \mathcal{F}_\text{reg}(\widehat{\text{res}}_u^{(p)}, \text{res}_u^{(p)}) ), \nonumber \\
\mathcal{L}_\text{res}^{(p)} &= \sum_{v \in \{y, h, w, l\}} \mathcal{F}_\text{reg}(\widehat{\text{res}}_v^{(p)}, \text{res}^{(p)}_v),  \label{reg_loss}\\
\mathcal{L}_\text{reg} &= \frac{1}{N_{\text{pos}}}\sum_{p \in \text{pos}}\left(\mathcal{L}_\text{bin}^{(p)} + \mathcal{L}_\text{res}^{(p)}\right) \nonumber
\end{align}
\vspace{-4mm}

\noindent
where $N_{\text{pos}}$ is the number of foreground points, $\widehat{\text{bin}}^{(p)}_u$ and $\widehat{\text{res}}^{(p)}_u$ are the predicted bin assignments and residuals of the foreground point $p$, $\text{bin}^{(p)}_u$ and $\text{res}^{(p)}_u$ are the ground-truth targets calculated as above,
$\mathcal{F}_\text{cls}$ denotes the cross-entropy classification loss, and $\mathcal{F}_\text{reg}$ denotes the smooth $L1$ loss.

To remove the redundant proposals, we conduct non-maximum suppression (NMS) based on the oriented IoU from bird's view to generate a small number of high-quality proposals. For training, we use 0.85 as the bird's view IoU threshold and after NMS we keep top 300 proposals for training the stage-2 sub-network. For inference,  we use oriented NMS with IoU threshold 0.8, and only top 100 proposals are kept for the refinement of stage-2 sub-network.

\subsection{Point cloud region pooling}
\label{ssec:pooling}

After obtaining 3D bounding box proposals, we aim at refining the box locations and orientations based on the previously generated box proposals. 
To learn more specific local features of each proposal, we propose to pool 3D points and their corresponding point features from stage-1 according to the location of each 3D proposal.

For each 3D box proposal, $\mathbf{b}_i=(x_i,y_i, z_i, h_i, w_i,$ $ l_i, \theta_i)$, we slightly enlarge it to create a new 3D box 
$\mathbf{b}_i^e=\left(x_i, y_i, z_i, h_i+\eta, w_i+\eta, l_i+\eta, \theta_i\right)$ to encode the additional information from its context, where $\eta$ is a constant value for enlarging the size of box. 

For each point $p = (x^{(p)}, y^{(p)}, z^{(p)})$, an inside/outside test is performed to determine whether the point $p$ is inside the enlarged bounding box proposal ${\bf b}_i^e$. If so, the point and its features would be kept for refining the box ${\bf b}_i$. 
The features associated with the inside point $p$ include its 3D point coordinates $(x^{(p)}, y^{(p)}, z^{(p)}) \in \mathbb{R}^3$, its laser reflection intensity $r^{(p)} \in \mathbb{R}$, 
its predicted segmentation mask $m^{(p)} \in \{0,1\}$ from stage-1,
and the $C$-dimensional learned point feature representation ${\bf f}^{(p)} \in \mathbb{R}^C$ from stage-1. 

We include the segmentation mask $m^{(p)}$ to differentiate the predicted foreground/background points within the enlarged box ${\bf b}_i^e$. 
The learned point feature ${\bf f}^{(p)}$ encodes valuable information via learning for segmentation and proposal generation therefore are also included. We eliminate the proposals that have no inside points in the following stage.

\subsection{Canonical 3D bounding box refinement}\label{subrefine}

As illustrated in Fig.~\ref{fig:framework}~(b), the pooled points and their associated features (see Sec.~\ref{ssec:pooling}) for each proposal are fed to our stage-2 sub-network for refining the 3D box locations as well as the foreground object confidence.

\begin{figure}[t]
	\begin{center}
		\includegraphics[width=0.98\linewidth,height=3.2cm]{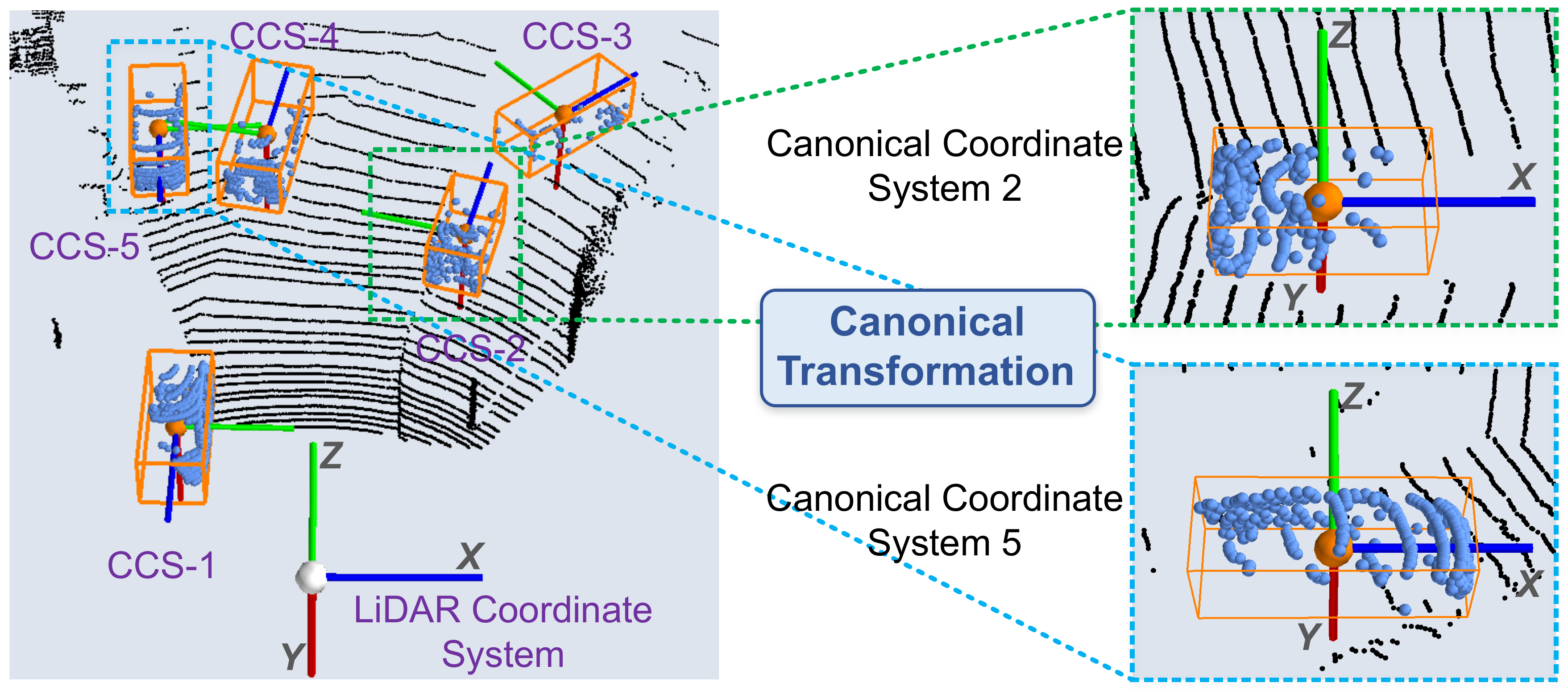}
	\end{center}
	\caption{Illustration of canonical transformation. The pooled points belonged to each proposal are transformed to the corresponding canonical coordinate system for better local spatial feature learning, where CCS denotes Canonical Coordinate System.}
	\label{fig:ct}
	\vspace{-3mm}
\end{figure}

\smallskip
\noindent
{\bf Canonical transformation.} To take advantages of our high-recall box proposals from stage-1 and to estimate only the residuals of the box parameters of proposals, we transform the pooled points belonging to each proposal to the canonical coordinate system of the corresponding 3D proposal. As shown in Fig.\ \ref{fig:ct}, the canonical coordinate system for one 3D proposal denotes that (1) the origin is located at the center of the box proposal; (2) the local $X'$ and $Z'$ axes are approximately parallel to the ground plane with $X'$ pointing towards the head direction of proposal and the other $Z'$ axis perpendicular to $X'$; (3) the $Y'$ axis remains the same as that of the LiDAR coordinate system. All pooled points' coordinates $p$ of the box proposal should be transformed to the canonical coordinate system as $\tilde{p}$ by proper rotation and translation. Using the proposed canonical coordinate system enables the box refinement stage to learn better local spatial features for each proposal.

\smallskip
\noindent
{\bf Feature learning for box proposal refinement.}
As we mentioned in Sec.~\ref{ssec:pooling}, the refinement sub-network combines both the transformed local spatial points (features) $\tilde{p}$ as well as their global semantic features ${\bf f}^{(p)}$ from stage-1 
for further box and confidence refinement.

Although the canonical transformation enables robust local spatial features learning, it inevitably loses depth information of each object. For instance, the far-away objects generally have much fewer points than nearby objects 
because of the fixed angular scanning resolution of the LiDAR sensors. 
To compensate for the lost depth information, we include the distance to the sensor, \ie, $d^{(p)}=\sqrt{(x^{(p)})^2 + (y^{(p)})^2 + (z^{(p)})^2}$, into the features of point $p$.

For each proposal, its associated points' local spatial features $\tilde{p}$ and the extra features $[r^{(p)}, m^{(p)}, d^{(p)}]$ are first concatenated and fed to several fully-connected layers to encode their local features 
to the same dimension of the global features ${\bf f}^{(p)}$.
Then the local features and global features are concatenated and fed into a network following the structure of \cite{qi2017pointnet++} to obtain a discriminative feature vector 
for the following confidence classification and box refinement.

\smallskip
\noindent
{\bf Losses for box proposal refinement.}
We adopt the similar bin-based regression losses for proposal refinement. 
A ground-truth box is assigned to a 3D box proposal for learning box refinement if their 3D IoU is greater than $0.55$. 
Both the 3D proposals and their corresponding 3D ground-truth boxes are transformed into the canonical coordinate systems, which means the 3D proposal $\mathbf{b}_i=(x_i, y_i, z_i, h_i, w_i, l_i, \theta_i)$ and 3D ground-truth box $\mathbf{b}_i^{\text{gt}}=(x_i^{\text{gt}}, y_i^{\text{gt}}, z_i^{\text{gt}}, h_i^{\text{gt}}, w_i^{\text{gt}}, l_i^{\text{gt}}, \theta_i^{\text{gt}})$ would be transformed to
\vspace{-2mm}
\begin{align}
\mathbf{\tilde{b}}_i &=(0, 0, 0, h_i, w_i, l_i, 0), \label{pcpool} \\
\mathbf{\tilde{b}}_i^{\text{gt}} &=(x_i^{\text{gt}} - x_i, y_i^{\text{gt}} - y_i, z_i^{\text{gt}} - z_i, h_i^{\text{gt}}, w_i^{\text{gt}}, l_i^{\text{gt}}, \theta_i^{\text{gt}} - \theta_i) \nonumber
\end{align}
\vspace{-5mm}

The training targets for the $i$th box proposal's center location,
$(\text{bin}_{\Delta x}^i, \text{bin}_{\Delta z}^i,$ $\text{res}_{\Delta x}^i, \text{res}_{\Delta z}^i, \text{res}_{\Delta y}^i)$, are set in the same way as Eq. (\ref{loc_loss}) except that we use smaller search range $\mathcal{S}$ for refining the locations of 3D proposals. We still directly regress size residual $(\text{res}_{\Delta h}^i, \text{res}_{\Delta w}^i, \text{res}_{\Delta l}^i)$ w.r.t. the average object size of each class in the training set since the pooled sparse points usually could not provide enough information of the proposal size $(h_i, w_i, l_i)$. 

For refining the orientation,
we assume that the angular difference w.r.t. the ground-truth orientation, $\theta_i^{\text{gt}} - \theta_i$, is within the range $[-\frac{\pi}{4}, \frac{\pi}{4}]$, based on the fact that
the 3D IoU between a proposal and their ground-truth box is at least $0.55$. 
Therefore, we divide $\frac{\pi}{2}$ into discrete bins with the bin size $\omega$ and 
predict the bin-based orientation targets as 
\vspace{-2mm}
\begin{align}\label{ort_loss}
\text{bin}_{\Delta\theta}^i  &= \floor*{\frac{\theta^{\text{gt}}_i - \theta_i + \frac{\pi}{4}}{\omega}},
\\
\text{res}_{\Delta\theta}^i  &= \frac{2}{\omega}\left( \theta^{\text{gt}}_i - \theta_i + \frac{\pi}{4} - \left(\text{bin}^i_{\Delta\theta} \cdot \omega + \frac{\omega}{2}\right)\right)\nonumber
\end{align}
Therefore, the overall loss for the stage-2 sub-network can be formulated as

\vspace{-2mm}
\begin{equation}\label{reg_loss2}
\begin{aligned}
\mathcal{L}_\text{refine} =& \frac{1}{|| \mathcal{B} ||} \sum_{i \in \mathcal{B}}
\mathcal{F}_\text{cls}(\text{prob}_i, \text{label}_i) \\
 &+ \frac{1}{||\mathcal{B}_{pos}||} \sum_{i \in \mathcal{B}_{pos}}
(\mathcal{\tilde{L}}_\text{bin}^{(i)} + \mathcal{\tilde{L}}_\text{res}^{(i)})
\end{aligned}
\end{equation}
where $\mathcal{B}$ is the set of 3D proposals from stage-1 and $\mathcal{B}_{pos}$ stores the positive proposals for regression, $\text{prob}_i$ is the estimated confidence of $\tilde{{\bf b}}_i$ and $\text{label}_i$ is the corresponding label, 
$\mathcal{F}_\text{cls}$ is the cross entropy loss to supervise the predicted confidence, 
$\mathcal{\tilde{L}}_\text{bin}^{(i)}$ and $\mathcal{\tilde{L}}_\text{res}^{(i)}$ are similar to $\mathcal{L}_\text{bin}^{(p)}$ and $\mathcal{L}_\text{res}^{(p)}$ in Eq. \eqref{reg_loss} with the new targets calculated by $\mathbf{\tilde{b}}_i$ and $\mathbf{\tilde{b}}_i^{\text{gt}}$ as above.

We finally apply oriented NMS with bird's view IoU threshold 0.01 to remove the overlapping bounding boxes and generate the 3D bounding boxes for detected objects.

\section{Experiments}
PointRCNN is evaluated on the challenging 3D object detection benchmark of KITTI dataset \cite{Geiger2012CVPR}. We first introduce the implementation details of PointRCNN in Sec.~\ref{imple}. In Sec.~\ref{compare}, we perform a comparison with state-of-the-art 3D detection methods. Finally, we conduct extensive ablation studies to analyze PointRCNN in Sec.~\ref{ablation}.

\begin{table*}
	\small 
	\begin{center}
		\scalebox{0.975}{
			\begin{tabular}{c|c||ccc|ccc|ccc}
				\hline
				\multirow{2}{*}{Method} & \multirow{2}{*}{Modality} & 			
				\multicolumn{3}{c|}{Car (IoU=0.7)} & \multicolumn{3}{c|}{Pedestrian (IoU=0.5)} & \multicolumn{3}{c}{Cyclist (IoU=0.5)} \\
				&&Easy & Moderate & Hard & Easy & Moderate & Hard & Easy & Moderate & Hard\\
				
				\hline\hline
				MV3D \cite{chen2017multi} & RGB + LiDAR & 71.09 & 62.35 & 55.12 & - & - & - & - & - & - \\
				UberATG-ContFuse \cite{liang2018deep} & RGB + LiDAR & 82.54 & 66.22 & 64.04 & - & - & - & - & - & - \\
				AVOD-FPN \cite{ku2017joint} & RGB + LiDAR & 81.94 & 71.88 & 66.38 & 50.80 & 42.81 & \textbf{40.88} & 64.00 & 52.18 & 46.61 \\			
				F-PointNet \cite{qi2017frustum} & RGB + LiDAR & 81.20 & 70.39 & 62.19 & \textbf{51.21} & \textbf{44.89} & 40.23 & 71.96 & 56.77 & 50.39\\			
				VoxelNet \cite{zhou2017voxelnet}& LiDAR & 77.47 & 65.11 & 57.73 & 39.48 & 33.69 & 31.51 & 61.22 & 48.36 & 44.37 \\
				SECOND \cite{yan2018second} & LiDAR & 83.13 & 73.66 & 66.20 & 51.07 & 42.56 & 37.29 & 70.51 & 53.85 & 46.90\\
				\hline 
				Ours & LiDAR & {\bf85.94} & {\bf75.76} & {\bf68.32} & 49.43 & 41.78 & 38.63 & {\bf73.93} & {\bf59.60} & {\bf53.59} \\
				\hline
			\end{tabular}
		}	
	\end{center}
	\caption{Performance comparison of 3D object detection with previous methods on KITTI \textit{test} split by submitting to official test server. The evaluation metric is Average Precision(AP) with IoU threshold 0.7 for car and 0.5 for pedestrian/cyclist.}
	\label{tab:test}
	\vspace{-3mm}
\end{table*}

\subsection{Implementation Details}\label{imple}
\smallskip
\noindent
{\bf Network Architecture.}
For each 3D point-cloud scene in the training set, we subsample 16,384 points from each scene as the inputs. For scenes with the number of points fewer than 16,384, we randomly repeat the points to obtain 16,384 points. For the stage-1 sub-network, we follow the network structure of \cite{qi2017pointnet++}, where four set-abstraction layers with multi-scale grouping are used to subsample points into groups with sizes 4096, 1024, 256, 64. Four feature propagation layers are then used to obtain the per-point feature vectors for segmentation and proposal generation.

For the box proposal refinement sub-network, we randomly sample 512 points from the pooled region of each proposal as the input of the refinement sub-network. Three set abstraction layers with single-scale grouping \cite{qi2017pointnet++} (with group sizes 128, 32, 1) are used to generate a single feature vector for object confidence classification and proposal location refinement.

\smallskip
\noindent
{\bf The training scheme.}
Here we report the training details of car category since it has the majority of samples in the KITTI dataset, and the proposed method could be extended to other categories (like pedestrian and  cyclist)  easily with little modifications of hyper parameters. 

For stage-1 sub-network, all points inside the 3D ground-truth boxes are considered as foreground points and others points are treated as background points. During training, we ignore background points near the object boundaries by enlarging the 3D ground-truth boxes by 0.2m on each side of object for robust segmentation since the 3D ground-truth boxes may have small variations. For the bin-based proposal generation, the hyper parameters are set as search range $S=3m$, bin size $\delta=0.5m$ and orientation bin number $n=12$.

To train the stage-2 sub-network, we randomly augment the 3D proposals with small variations to increase the diversity of proposals. For training the box classification head, a proposal is considered as positive if its maximum 3D IoU with ground-truth boxes is above 0.6, and is treated as negative if its maximum 3D IoU is below 0.45. We use 3D IoU 0.55 as the minimum threshold of proposals for the training of box regression head.
For the bin-based proposal refinement, search range is $S=1.5m$, localization bin size is $\delta=0.5m$ and orientation bin size is $\omega=\ang{10}$. The context length of point cloud pooling is $\eta=1.0m$. 

The two stage sub-networks of PointRCNN are trained separately. The stage-1 sub-network is trained for 200 epochs with batch size 16 and learning rate 0.002, 
while the stage-2 sub-network is trained for 50 epochs with batch size 256 and learning rate 0.002. During training, we conduct data augmentation of random flip, scaling with a scale factor sampled from [0.95, 1.05] and rotation around vertical $Y$ axis between [-10, 10] degrees. Inspired by \cite{yan2018second}, to simulate objects with various environments, we also put several new ground-truth boxes and their inside points from other scenes to the same locations of current training scene by randomly selecting non-overlapping boxes, and this augmentation is denoted as GT-AUG in the following sections.

\subsection{3D Object Detection on KITTI}\label{compare}
The 3D object detection benchmark of KITTI contains 7481 training samples and 7518 testing samples (\textit{test} split). We follow the frequently used train/val split mentioned in \cite{chen2017multi} to divide the training samples into \textit{train} split (3712 samples) and \textit{val} split (3769 samples). We compare PointRCNN with state-of-the-art methods of 3D object detection on both \textit{val} split and \textit{test} split of KITTI dataset. All the models are trained on \textit{train} split and evaluated on \textit{test} split and \textit{val} split.

\noindent
{\bf Evaluation of 3D object detection.}\quad 
We evaluate our method on the 3D detection benchmark of the KITTI test server, and the results are shown in Tab.~\ref{tab:test}. 
For the 3D detection of car and cyclist, our method outperforms previous state-of-the-art methods with remarkable margins on all three difficulties and ranks first on the KITTI test board among all published works at the time of submission. Although most of the previous methods use both RGB image and point cloud as input, our method achieves better performance with an efficient architecture by using only the point cloud as input.
For the pedestrian detection, compared with previous LiDAR-only methods, our method achieves better or comparable results, but it performs slightly worse than the methods with multiple sensors. 
We consider it is due to the fact that our method only uses sparse point cloud as input but pedestrians have small size and image could capture more details of pedestrians than point cloud to help 3D detection.

For the most important car category, we also report the performance of 3D detection result on the \textit{val} split as shown in Tab.~\ref{tab:val}. Our method outperforms previous stage-of-the-art methods with large margins on the \textit{val} split. Especially in the hard difficulty, our method has 8.28\% AP improvement than the previous best AP, which demonstrates the effectiveness of the proposed PointRCNN. 

\begin{table}
	\small 
	\begin{center}
		\scalebox{1.0}{
			\begin{tabular}{c|ccc}
				\hline
				\multirow{2}{*}{Method} & 			
				\multicolumn{3}{c}{AP(IoU=0.7)}\\
				& Easy & Moderate & Hard \\
				
				\hline
				MV3D \cite{chen2017multi} & 71.29 & 62.68 & 56.56 \\
				VoxelNet \cite{zhou2017voxelnet}& 81.98 & 65.46 & 62.85 \\
				SECOND \cite{yan2018second} & 87.43 & 76.48 & 69.10 \\
				AVOD-FPN \cite{ku2017joint} & 84.41 & 74.44 & 68.65 \\			
				F-PointNet \cite{qi2017frustum} & 83.76 & 70.92 & 63.65 \\			
				\hline 
				Ours (no GT-AUG) & 88.45 & 77.67 & 76.30 \\
				Ours & {\bf88.88} & {\bf 78.63} & {\bf 77.38} \\
				\hline
			\end{tabular}
		}
	\end{center}
	\caption{Performance comparison of 3D object detection with previous methods on the car class of KITTI \emph{val} split set.}
	\label{tab:val}
	\vspace{-0.5cm}
\end{table}

\noindent
{\bf Evaluation of 3D proposal generation.}\quad 
The performance of our bottom-up proposal generation network is evaluated by calculating the recall of 3D bounding box with various number of proposals and 3D IoU threshold. As shown in Tab.~\ref{tab:recall}, our method (without GT-AUG) achieved significantly higher recall than previous methods. With only 50 proposals, our method obtains \textbf{96.01\%} recall at IoU threshold 0.5 on the moderate difficulty of car class, which outperforms recall 91\% of AVOD \cite{ku2017joint} by 5.01\% at the same number of proposals, note that the latter method uses both 2D image and point cloud for proposal generation while we only use point cloud as input. When using 300 proposals, our method further achieves \textbf{98.21\%} recall at IoU threshold 0.5. It is meaningless to increase the number of proposals since our method already obtained high recall at IoU threshold 0.5. In contrast, as shown in Tab.~\ref{tab:recall}, we report the recall of 3D bounding box at IoU threshold 0.7 for reference. With 300 proposals, our method achieves \textbf{82.29\%} recall at IoU threshold 0.7. Although the recall of proposals are loosely \cite{hosang2016makes, girshick2015fast} related to the final 3D object detection performance, the outstanding recall still suggests the robustness and accuracy of our bottom-up proposal generation network.

\begin{table}
	\small 
	\begin{center}
		\scalebox{0.95}{
			\begin{tabular}{c|ccc|c}
				\hline 
				\multirow{2}{*}{RoIs \#} & 			
				\multicolumn{3}{c|}{Recall(IoU=0.5)} & 
				\multicolumn{1}{c}{Recall(IoU=0.7)}\\
				& MV3D & AVOD & Ours & Ours\\
				\hline
				10 & - &  86.00 & {\bf86.66} & 29.87\\
				20 & - & - & {\bf91.83} & 32.55 \\
				30 & - & - & {\bf93.31} & 32.76 \\
				40 & - & - & {\bf95.55} & 40.04 \\
				50 & - & 91.00 & {\bf96.01} & 40.28 \\
				100 & - & - & {\bf96.79} & 74.81 \\
				200 & - & - & {\bf98.03} & 76.29 \\ 
				300 & 91.00 & - & {\bf98.21} & 82.29 \\
				\hline 
			\end{tabular}
		}
	\end{center}
	\caption{Recall of proposal generation network with different number of RoIs and 3D IoU threshold for the car class on the \emph{val} split at moderate difficulty. Note that only MV3D \cite{chen2017multi} and AVOD \cite{ku2017joint} of previous methods reported the number of recall.}
	\label{tab:recall}
	\vspace{-3mm}
\end{table}

\subsection{Ablation Study}\label{ablation}
In this section, we conduct extensive ablation experiments to analyze the effectiveness of different components of PointRCNN. All experiments are trained on the \textit{train} split without GT-AUG and evaluated on the \textit{val} split with the car class\footnote{The KITTI test server only allows 3 submissions in every 30 days. All previous methods conducted ablation studies on the validation set.}.

\smallskip
\noindent
{\bf Different inputs for the refinement sub-network.} 
As mentioned in Sec.~\ref{subrefine}, the inputs of the refinement sub-network consist of the canonically transformed coordinates and pooled features of each pooled point. 

We analyze the effects of each type of features to the refinement sub-network by removing one and keeping all other parts unchanged. All experiments share the same fixed stage-1 sub-network for fair comparison. The results are shown in Tab.~\ref{tab:input_comb}.
Without the proposed canonical transformation, the performance of the refinement sub-network dropped significantly, which shows the transformation into a canonical coordinate system greatly eliminates much rotation and location variations and improve the efficiency of feature learning for the stage-2.
We also see that removing the stage-1 features ${\bf f}^{(p)}$ learned from point cloud segmentation and proposal generation decreases the mAP by 2.71\% on the moderate difficulty, which demonstrates the advantages of learning for semantic segmentation in the first stage.
Tab.~\ref{tab:input_comb} also shows that the camera depth information $d^{(p)}$ and segmentation mask $m^{(p)}$ for 3D points $p$ contribute slightly to the final performance, since the camera depth completes the distance information which is eliminated during the canonical transformation and the segmentation mask indicates the foreground points in the pooled regions. 

\begin{table}
	\small 
	\begin{center}
		\scalebox{1.0}{
			\begin{tabular}{c|c|c|c|ccc}
				\hline		
				\tabincell{c}{CT}
				& \tabincell{c}{RPN \\features} 
				& \tabincell{c}{camera\\depth} 
				& \tabincell{c}{seg.\\ mask}
				& $AP_E$ & $AP_M$ & $AP_H$ \\
				\hline
				$\times$ & \checkmark & \checkmark & \checkmark & 7.64 & 13.68 & 13.94\\
				\checkmark & $\times$  & \checkmark & \checkmark & 84.75 & 74.96 & 74.29 \\
				\checkmark & \checkmark & $\times$  & \checkmark & 87.34 & 76.79 & 75.46 \\
				\checkmark & \checkmark & \checkmark & $\times$  & 86.25 & 76.64 & 75.86 \\
				\checkmark & \checkmark & \checkmark & \checkmark  & {\bf 88.45} & {\bf77.67} &{\bf 76.30} \\
				\hline
			\end{tabular}
		}
	\end{center}
	\caption{Performance for different input combinations of refinement network. $AP_E$, $AP_M$, $AP_H$ denote the average precision for easy, moderate, hard difficulty on KITTI \textit{val} split, respectively. CT denotes canonical transformation.}
	\label{tab:input_comb}
	\vspace{-0.1cm}
\end{table}

\smallskip
\noindent
{\bf Context-aware point cloud pooling. }\quad
In Sec.~\ref{ssec:pooling}, we introduce enlarging the proposal boxes ${\bf b}_i$ by a margin $\eta$ to create ${\bf b}_i^e$ to pool more contextual points for each proposal's confidence estimation and location regression. Tab.~\ref{tab:pool} shows the effects of different pooled context widths $\eta$. $\eta = 1.0m$ results in the best performance in our proposed framework. We notice that when no contextual information is pooled, the accuracies, especially those at the hard difficulty, drops significantly. The difficult cases often have fewer points in the proposals since the object might be occluded or far away from the sensor, which needs more context information for classification and proposal refinement. As shown in Tab.~\ref{tab:pool}, too large $\eta$ also leads to performance drops since the pooled region of current proposals may include noisy foreground points of other objects.

\begin{table}
	\small 
	\begin{center}
		\scalebox{1.0}{
			\begin{tabular}{c|ccc}
				\hline 
				$\eta$~(context width) & $AP_E$ & $AP_M$ & $AP_H$ \\
				\hline 
				no context & 86.65 & 75.68 & 68.92 \\
				0.5m & 87.87 & 77.12 & 75.61 \\
				0.8m & 88.27 & 77.40 & 76.07 \\
				1.0m & {\bf 88.45} & {\bf77.67} &{\bf 76.30}\\
				1.5m & 86.82 & 76.87 & 75.88 \\
				2.0m & 86.47 & 76.61 & 75.53\\
				\hline 
			\end{tabular}
		}
	\end{center}
	\caption{Performance of adopting different context width $\eta$ of context-aware point cloud pooling.}
	\label{tab:pool}
	\vspace{-0.4cm}
\end{table}

\smallskip
\noindent
{\bf Losses of 3D bounding box regression. }\quad 
In Sec.~\ref{sec:rpn}, we propose the bin-based localization losses for generating 3D box proposals. In this part, we evaluate the performances when using different types of 3D box regression loss for our stage-1 sub-network, which include the residual-based loss (RB-loss) \cite{zhou2017voxelnet}, residual-cos-based loss (RCB-loss), corner loss (CN-loss) \cite{chen2017multi, ku2017joint}, partial-bin-based loss (PBB-loss) \cite{qi2017frustum}, and our full bin-based loss (BB-loss). Here the residual-cos-based loss encodes $\Delta \theta$ of residual-based loss by $(cos(\Delta\theta), sin(\Delta\theta))$ to eliminate the ambiguity of angle regression.

\begin{figure}[t]
	\begin{center}
		\includegraphics[width=0.98\linewidth, height=5.6cm]{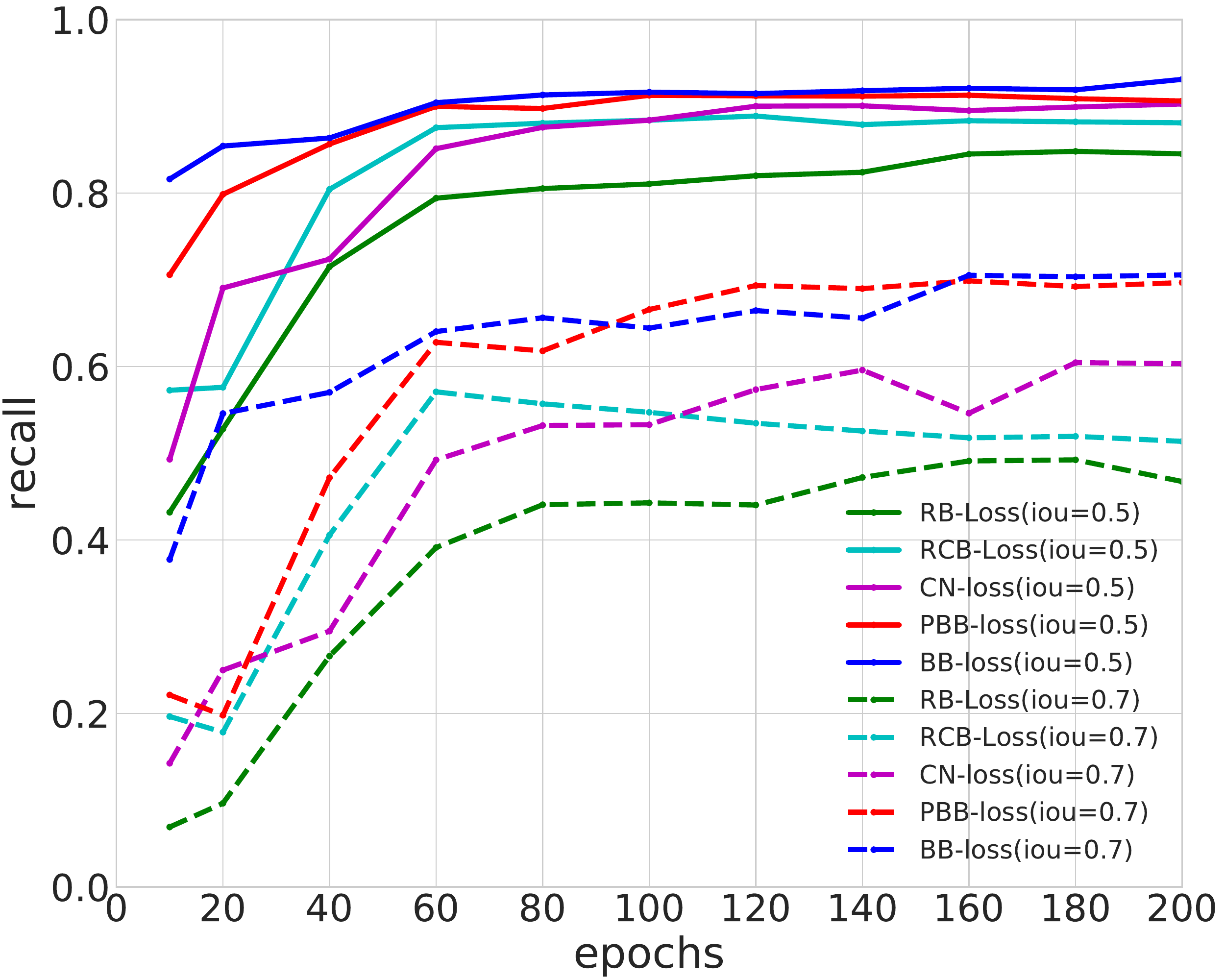}
	\end{center}
	\vspace{-3mm}
	\caption{Recall curves of applying different bounding box regression loss function.}
	\label{fig:loss_recall}
	\vspace{-0.3cm}
\end{figure}

The final recall (IoU thresholds 0.5 and 0.7) with 100 proposals from stage-1 are used as the evaluation metric, which are shown in Fig.~\ref{fig:loss_recall}. The plot reveals the effectiveness of our full bin-based 3D bounding box regression loss.
Specifically, stage-1 sub-network with our full bin-based loss function achieves higher recall and converges much faster than all other loss functions, which benefits from constraining the targets, especially  the localization, with prior knowledge. The partial-bin-based loss achieves similar recall but the convergence speed is much slower than ours. Both full and partial bin-based loss have significantly higher recall than other loss functions, especially at IoU threshold 0.7. The improved residual-cos-based loss also obtains better recall than residual-based loss by improving the angle regression targets.

\subsection{Qualitative Results}
\begin{figure*}
	\centering
	\small
	\scalebox{0.99}{
		\begin{tabular}{@{\hspace{0.0mm}}c@{\hspace{1.0mm}}c@{\hspace{1.0mm}}c@{\hspace{1.0mm}}c}
			\includegraphics[width=0.25\linewidth]{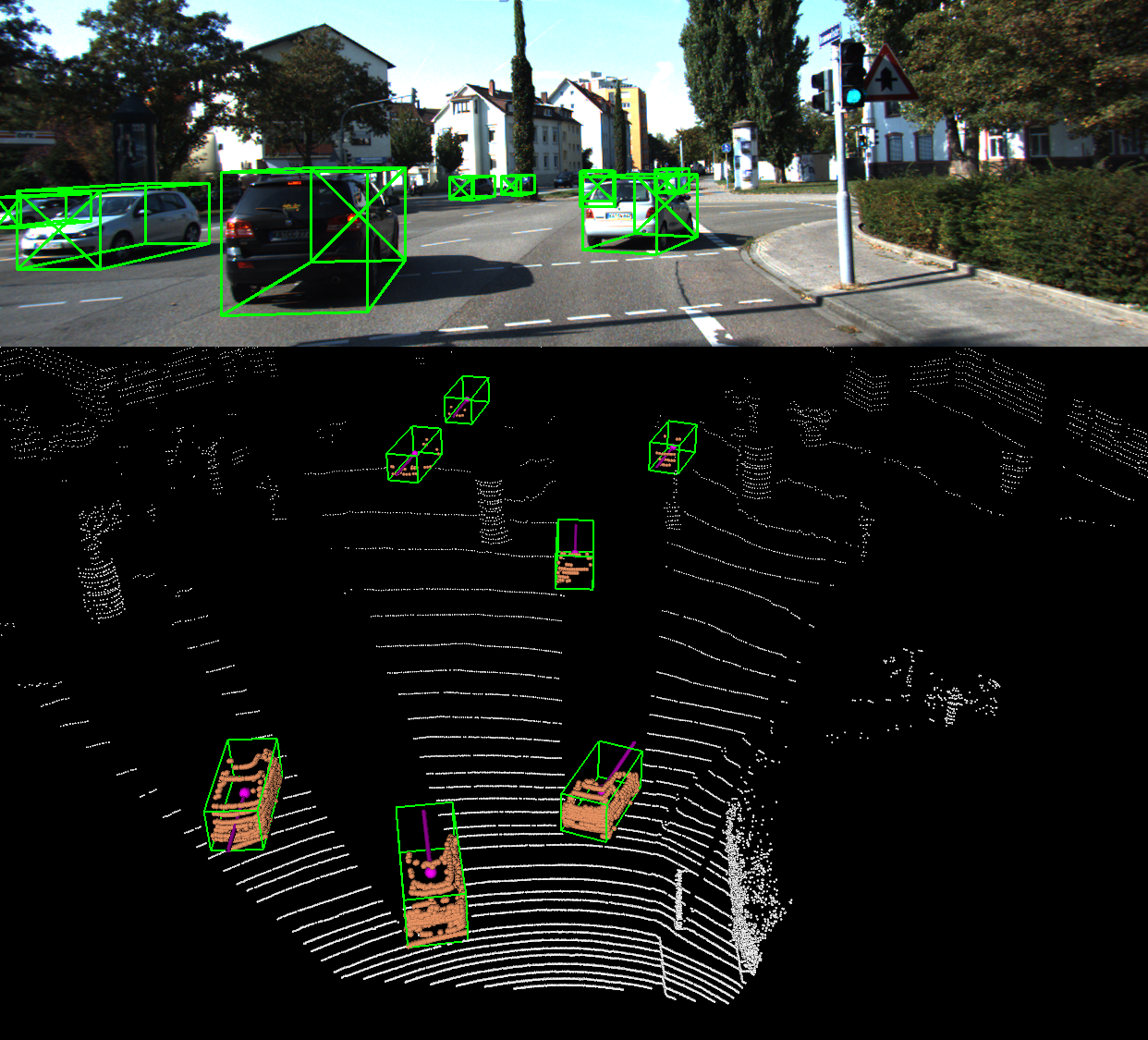}&
			\includegraphics[width=0.25\linewidth]{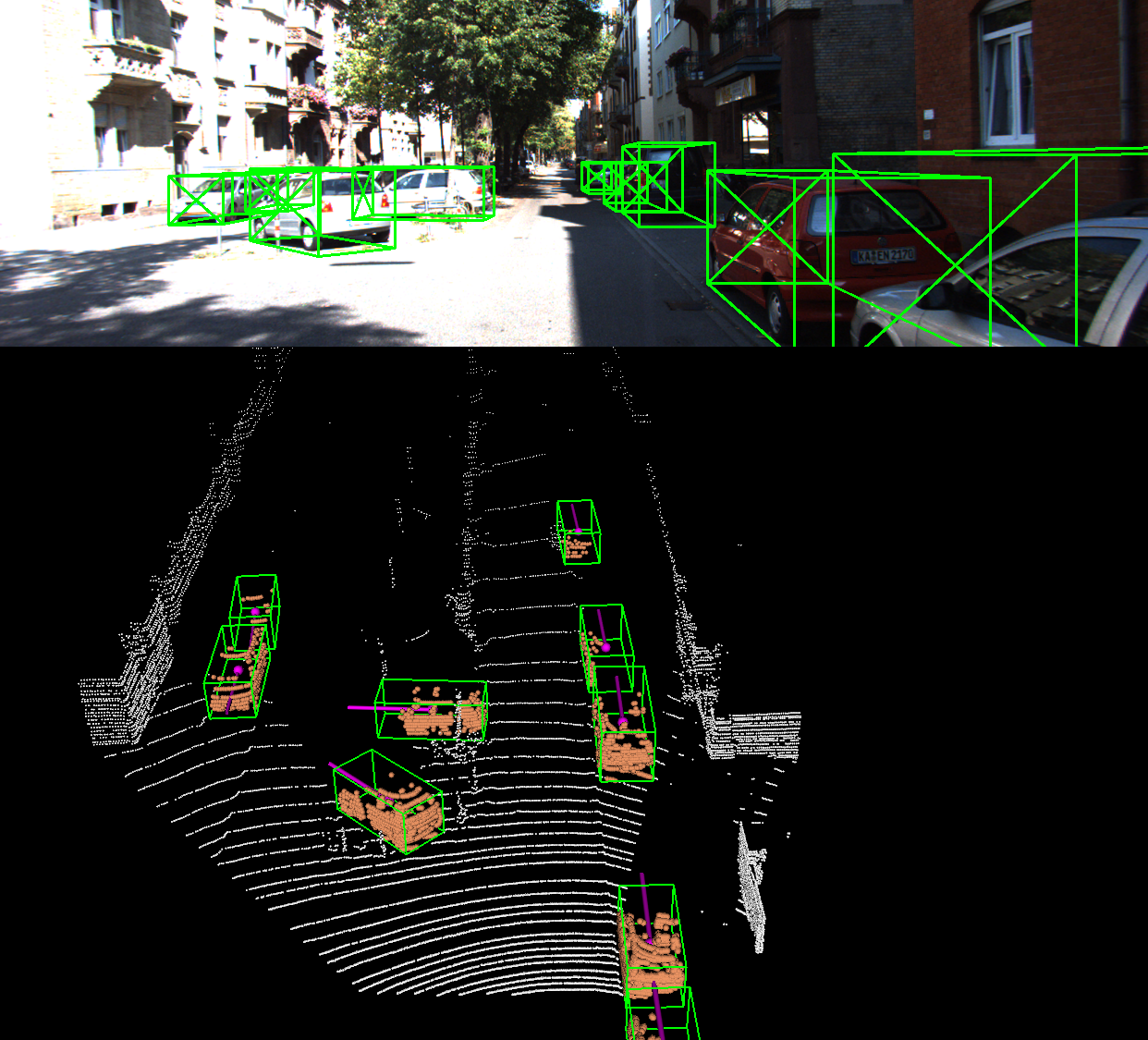}&
			\includegraphics[width=0.25\linewidth]{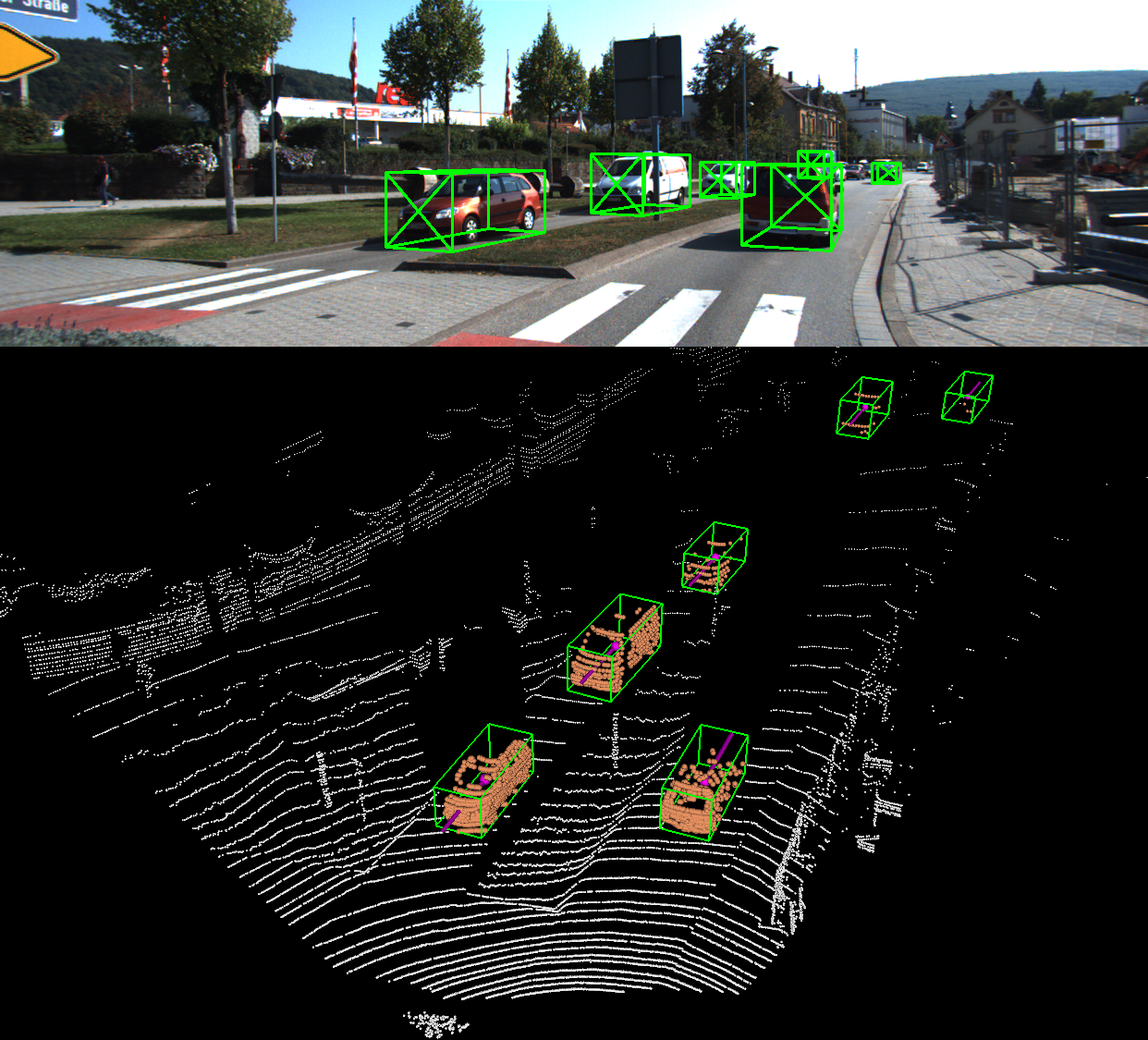}&
			\includegraphics[width=0.25\linewidth]{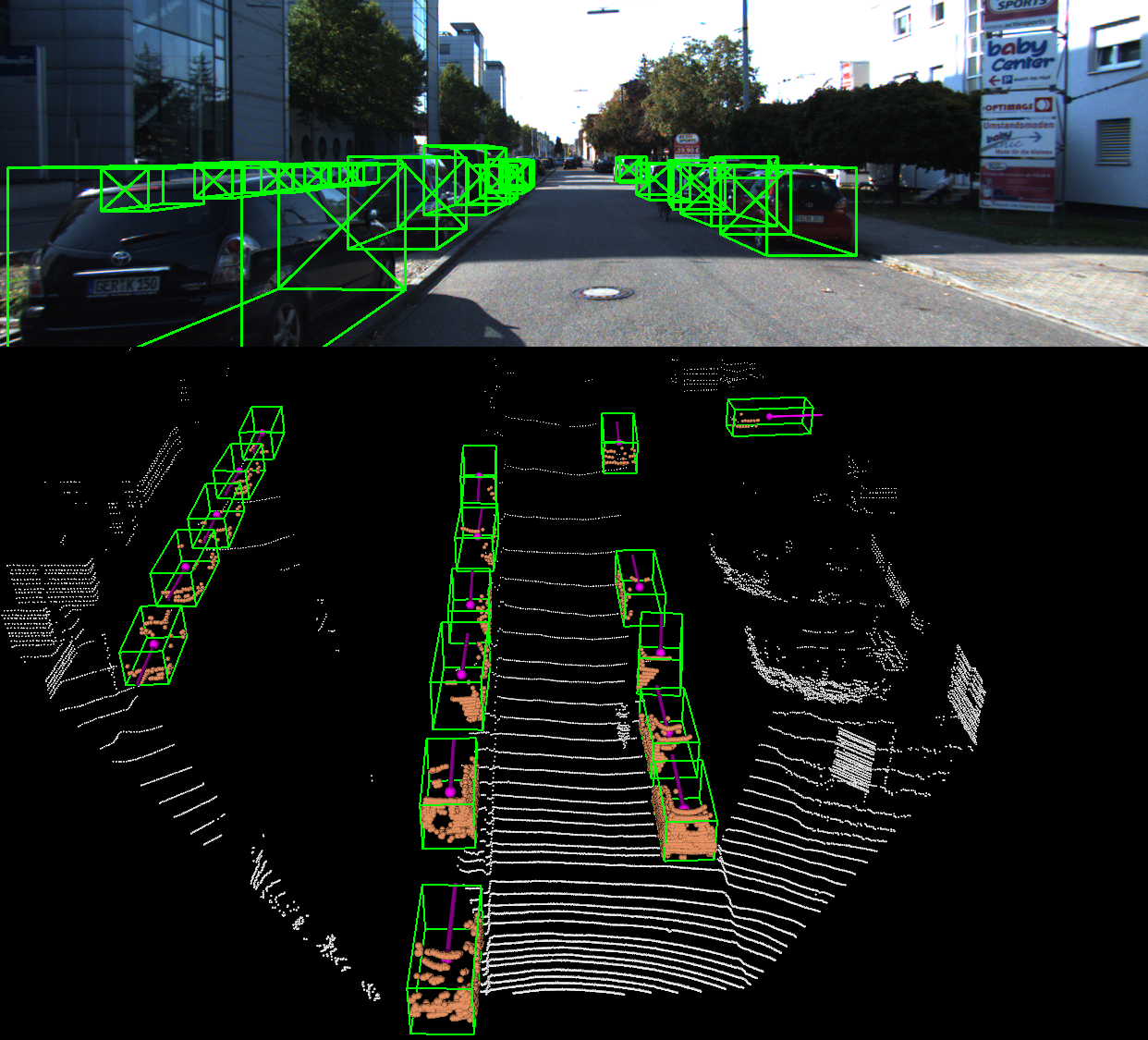}\\
			\includegraphics[width=0.25\linewidth]{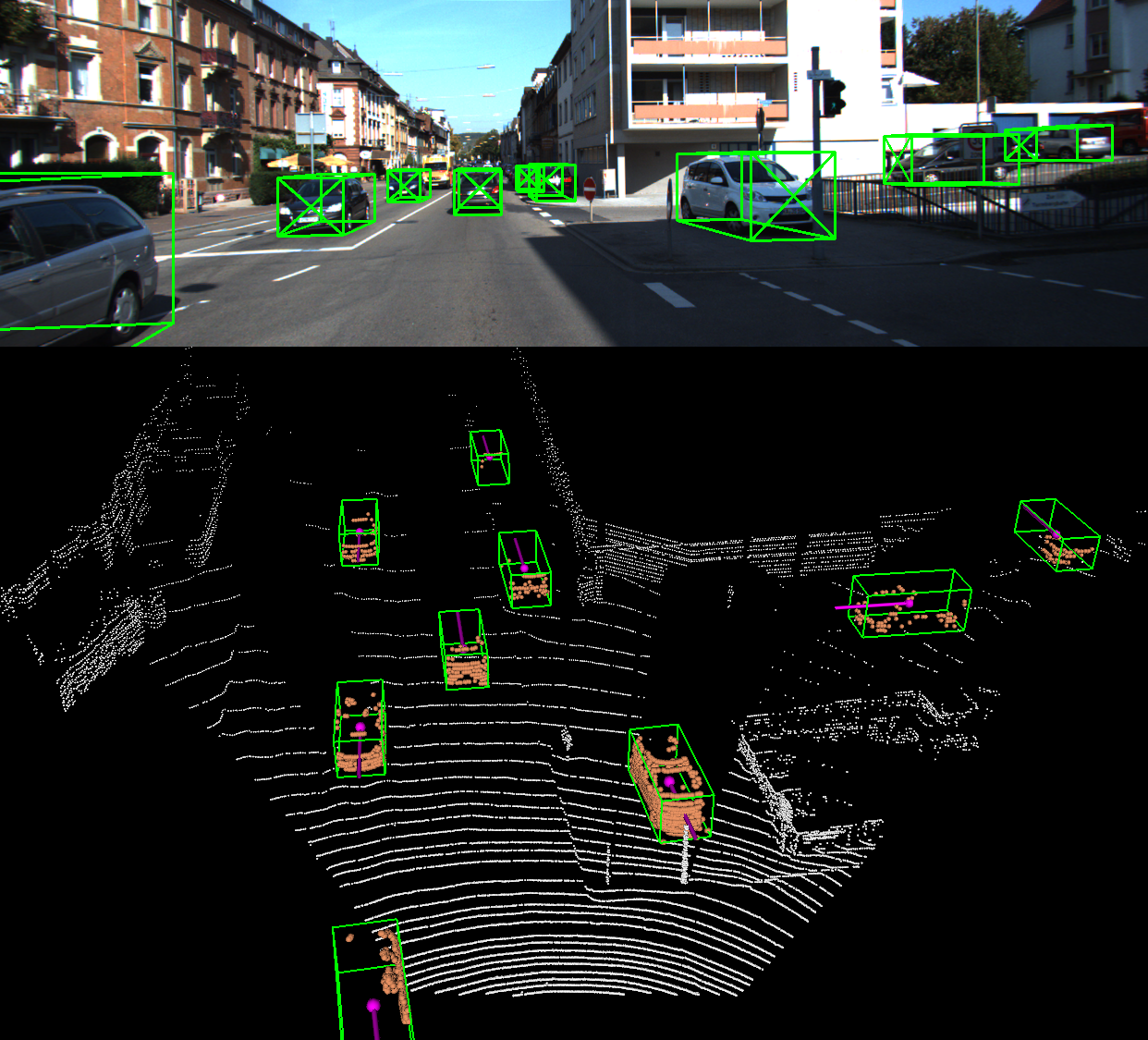}&
			\includegraphics[width=0.25\linewidth]{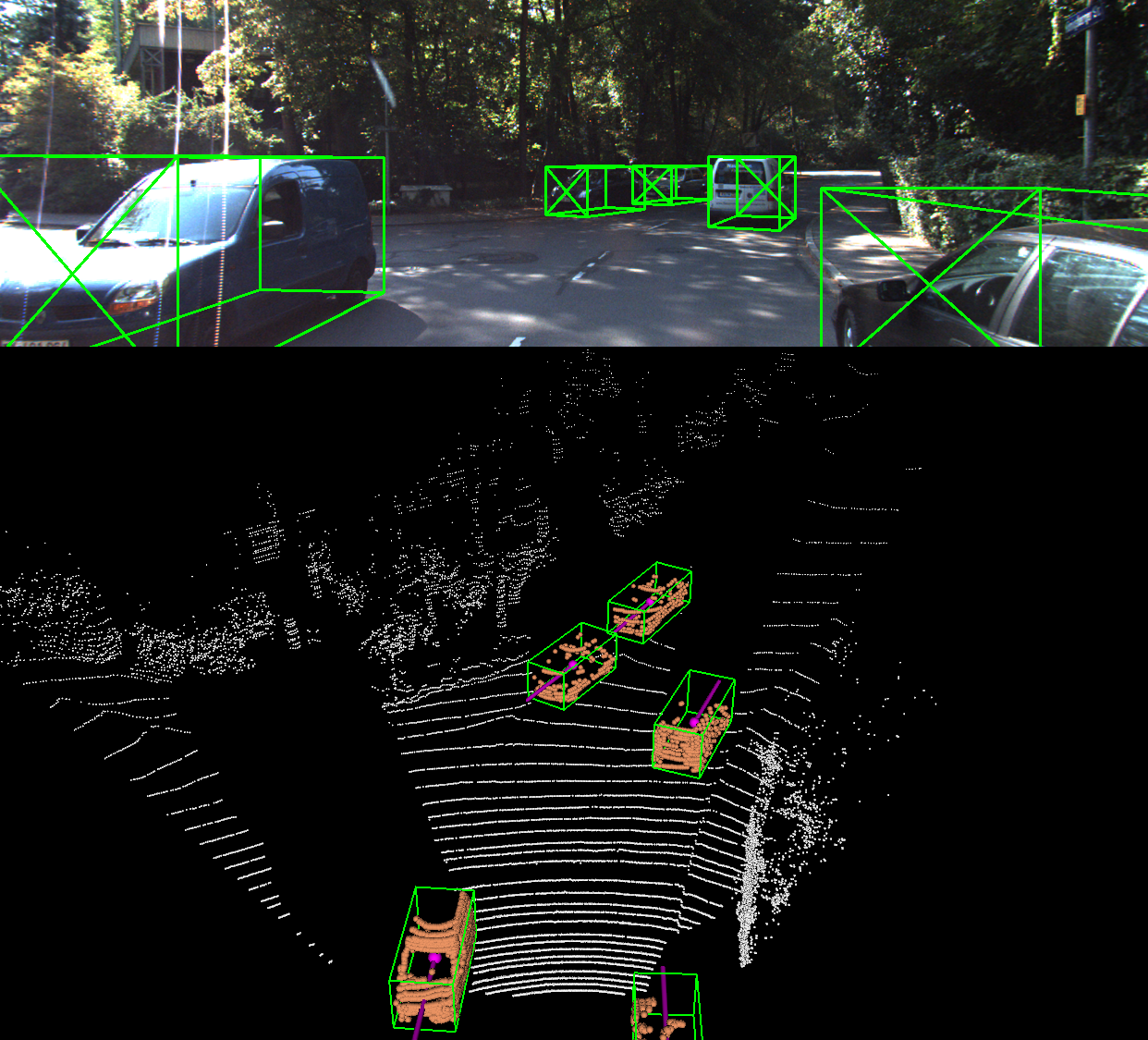}&
			\includegraphics[width=0.25\linewidth]{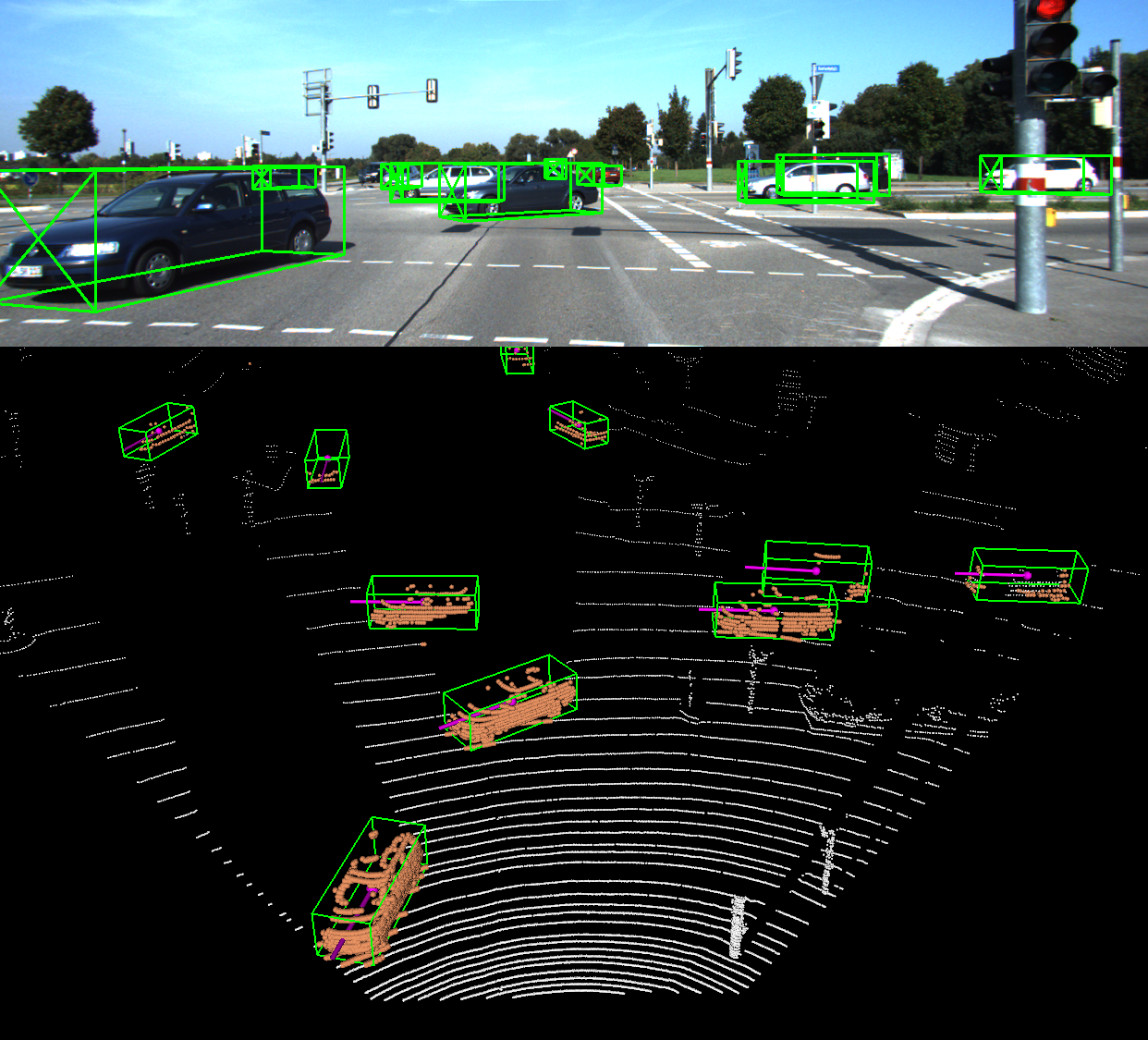}&
			\includegraphics[width=0.25\linewidth]{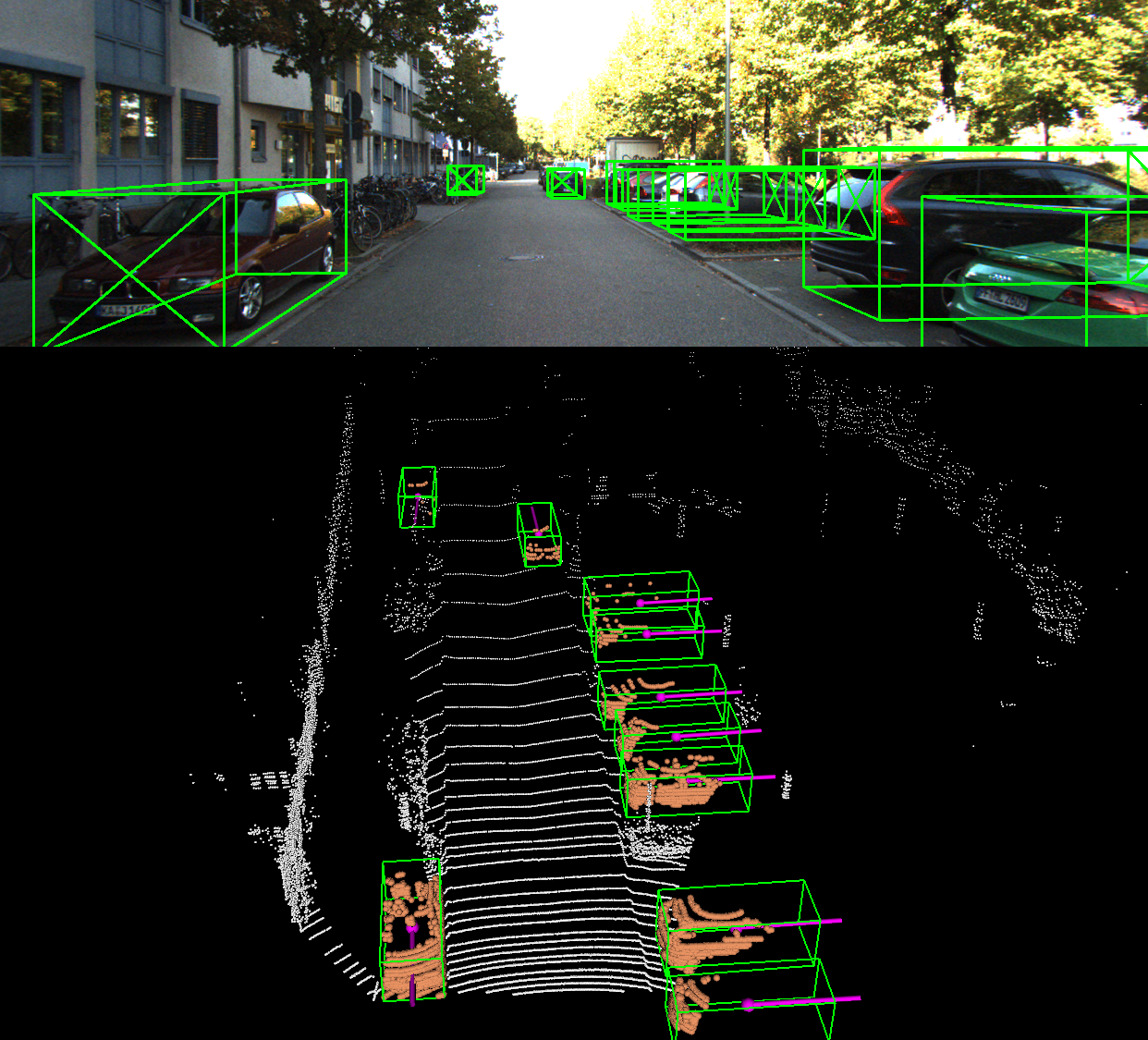}\\
	\end{tabular}}
	\caption{Qualitative results of PointRCNN on the KITTI \emph{test} split. For each sample, the upper part is the image and the lower part is a representative view of the corresponding point cloud. The detected objects are shown with green 3D bounding boxes, and the orientation (driving direction) of each object is specified by a X in the upper part and a red tube in the lower part. (Best viewed with zoom-in.)
	}
	\label{fig:test_vis}
	\vspace{-0.3cm}
\end{figure*}

Fig.~\ref{fig:test_vis} shows some qualitative results of our proposed PointRCNN on the \emph{test} split of KITTI \cite{Geiger2012CVPR} dataset. 
Note that the image is just for better visualization and our PointRCNN takes only the point cloud as input to generation 3D detection results.

\section{Conclusion}
We have presented PointRCNN, a novel 3D object detector for detecting 3D objects from raw point cloud. The proposed stage-1 network directly generates 3D proposals from point cloud in a bottom-up manner, which achieves significantly higher recall than previous proposal generation methods. The stage-2 network refines the proposals in the canonical coordinate by combining semantic features and local spatial features. Moreover, the newly proposed bin-based loss has demonstrated its efficiency and effectiveness for 3D bounding box regression. The experiments show that PointRCNN outperforms previous state-of-the-art methods with remarkable margins on the challenging 3D detection benchmark of KITTI dataset.

\section*{Acknowledgment}
This work is supported in part by SenseTime Group Limited, in part by the General Research Fund through the Research Grants Council of Hong Kong under Grants CUHK14202217, CUHK14203118, CUHK14205615, CUHK14207814, CUHK14213616, CUHK14208417, CUHK14239816, and in part by CUHK Direct Grant. 

{\small
\bibliographystyle{ieee}
\bibliography{egbib}

\begin{thebibliography}{10}\itemsep=-1pt

\bibitem{chabot2017deep}
Florian Chabot, Mohamed Chaouch, Jaonary Rabarisoa, C{\'e}line Teuli{\`e}re,
  and Thierry Chateau.
\newblock Deep manta: A coarse-to-fine many-task network for joint 2d and 3d
  vehicle analysis from monocular image.
\newblock In {\em Proc. IEEE Conf. Comput. Vis. Pattern Recognit.(CVPR)}, pages
  2040--2049, 2017.

\bibitem{chen2016monocular}
Xiaozhi Chen, Kaustav Kundu, Ziyu Zhang, Huimin Ma, Sanja Fidler, and Raquel
  Urtasun.
\newblock Monocular 3d object detection for autonomous driving.
\newblock In {\em Proceedings of the IEEE Conference on Computer Vision and
  Pattern Recognition}, pages 2147--2156, 2016.

\bibitem{chen20153d}
Xiaozhi Chen, Kaustav Kundu, Yukun Zhu, Andrew~G Berneshawi, Huimin Ma, Sanja
  Fidler, and Raquel Urtasun.
\newblock 3d object proposals for accurate object class detection.
\newblock In {\em Advances in Neural Information Processing Systems}, pages
  424--432, 2015.

\bibitem{chen2017multi}
Xiaozhi Chen, Huimin Ma, Ji Wan, Bo Li, and Tian Xia.
\newblock Multi-view 3d object detection network for autonomous driving.
\newblock In {\em Proceedings of the IEEE Conference on Computer Vision and
  Pattern Recognition}, pages 1907--1915, 2017.

\bibitem{dai2015boxsup}
Jifeng Dai, Kaiming He, and Jian Sun.
\newblock Boxsup: Exploiting bounding boxes to supervise convolutional networks
  for semantic segmentation.
\newblock In {\em Proceedings of the IEEE International Conference on Computer
  Vision}, pages 1635--1643, 2015.

\bibitem{dai2016instance}
Jifeng Dai, Kaiming He, and Jian Sun.
\newblock Instance-aware semantic segmentation via multi-task network cascades.
\newblock In {\em Proceedings of the IEEE Conference on Computer Vision and
  Pattern Recognition}, pages 3150--3158, 2016.

\bibitem{Geiger2012CVPR}
Andreas Geiger, Philip Lenz, and Raquel Urtasun.
\newblock Are we ready for autonomous driving? the kitti vision benchmark
  suite.
\newblock In {\em Conference on Computer Vision and Pattern Recognition
  (CVPR)}, 2012.

\bibitem{girshick2015fast}
Ross Girshick.
\newblock Fast r-cnn.
\newblock In {\em Proceedings of the IEEE international conference on computer
  vision}, pages 1440--1448, 2015.

\bibitem{3DSemanticSegmentationWithSubmanifoldSparseConvNet}
Benjamin Graham, Martin Engelcke, and Laurens van~der Maaten.
\newblock 3d semantic segmentation with submanifold sparse convolutional
  networks.
\newblock {\em CVPR}, 2018.

\bibitem{he2017mask}
Kaiming He, Georgia Gkioxari, Piotr Doll{\'a}r, and Ross Girshick.
\newblock Mask r-cnn.
\newblock In {\em Computer Vision (ICCV), 2017 IEEE International Conference
  on}, pages 2980--2988. IEEE, 2017.

\bibitem{hosang2016makes}
Jan Hosang, Rodrigo Benenson, Piotr Doll{\'a}r, and Bernt Schiele.
\newblock What makes for effective detection proposals?
\newblock {\em IEEE transactions on pattern analysis and machine intelligence},
  38(4):814--830, 2016.

\bibitem{huang2018recurrent}
Qiangui Huang, Weiyue Wang, and Ulrich Neumann.
\newblock Recurrent slice networks for 3d segmentation of point clouds.
\newblock In {\em Proceedings of the IEEE Conference on Computer Vision and
  Pattern Recognition}, pages 2626--2635, 2018.

\bibitem{jiang2018pointsift}
Mingyang Jiang, Yiran Wu, and Cewu Lu.
\newblock Pointsift: {A} sift-like network module for 3d point cloud semantic
  segmentation.
\newblock {\em CoRR}, abs/1807.00652, 2018.

\bibitem{ku2017joint}
Jason Ku, Melissa Mozifian, Jungwook Lee, Ali Harakeh, and Steven~Lake
  Waslander.
\newblock Joint 3d proposal generation and object detection from view
  aggregation.
\newblock {\em CoRR}, abs/1712.02294, 2017.

\bibitem{li2019gs3d}
Buyu Li, Wanli Ouyang, Lu Sheng, Xingyu Zeng, and Xiaogang Wang.
\newblock Gs3d: An efficient 3d object detection framework for autonomous
  driving.
\newblock 2019.

\bibitem{li2019_internet}
Hongyang Li, Bo Dai, Shaoshuai Shi, Wanli Ouyang, and Xiaogang Wang.
\newblock {Feature Intertwiner for Object Detection}.
\newblock In {\em ICLR}, 2019.

\bibitem{liang2018deep}
Ming Liang, Bin Yang, Shenlong Wang, and Raquel Urtasun.
\newblock Deep continuous fusion for multi-sensor 3d object detection.
\newblock In {\em Proceedings of the European Conference on Computer Vision
  (ECCV)}, pages 641--656, 2018.

\bibitem{lin2017feature}
Tsung-Yi Lin, Piotr Doll{\'a}r, Ross Girshick, Kaiming He, Bharath Hariharan,
  and Serge Belongie.
\newblock Feature pyramid networks for object detection.
\newblock In {\em Proceedings of the IEEE Conference on Computer Vision and
  Pattern Recognition}, pages 2117--2125, 2017.

\bibitem{lin2018focal}
Tsung-Yi Lin, Priyal Goyal, Ross Girshick, Kaiming He, and Piotr Doll{\'a}r.
\newblock Focal loss for dense object detection.
\newblock {\em IEEE transactions on pattern analysis and machine intelligence},
  2018.

\bibitem{liu2017sgn}
Shu Liu, Jiaya Jia, Sanja Fidler, and Raquel Urtasun.
\newblock Sgn: Sequential grouping networks for instance segmentation.
\newblock In {\em The IEEE International Conference on Computer Vision (ICCV)},
  2017.

\bibitem{liu2016ssd}
Wei Liu, Dragomir Anguelov, Dumitru Erhan, Christian Szegedy, Scott Reed,
  Cheng-Yang Fu, and Alexander~C Berg.
\newblock Ssd: Single shot multibox detector.
\newblock In {\em European conference on computer vision}, pages 21--37.
  Springer, 2016.

\bibitem{maturana2015voxnet}
Daniel Maturana and Sebastian Scherer.
\newblock Voxnet: A 3d convolutional neural network for real-time object
  recognition.
\newblock In {\em Intelligent Robots and Systems (IROS), 2015 IEEE/RSJ
  International Conference on}, pages 922--928. IEEE, 2015.

\bibitem{mottaghi2015coarse}
Roozbeh Mottaghi, Yu Xiang, and Silvio Savarese.
\newblock A coarse-to-fine model for 3d pose estimation and sub-category
  recognition.
\newblock In {\em Proceedings of the IEEE Conference on Computer Vision and
  Pattern Recognition}, pages 418--426, 2015.

\bibitem{mousavian20173d}
Arsalan Mousavian, Dragomir Anguelov, John Flynn, and Jana Ko{\v{s}}eck{\'a}.
\newblock 3d bounding box estimation using deep learning and geometry.
\newblock In {\em Computer Vision and Pattern Recognition (CVPR), 2017 IEEE
  Conference on}, pages 5632--5640. IEEE, 2017.

\bibitem{qi2017frustum}
Charles~Ruizhongtai Qi, Wei Liu, Chenxia Wu, Hao Su, and Leonidas~J. Guibas.
\newblock Frustum pointnets for 3d object detection from {RGB-D} data.
\newblock {\em CoRR}, abs/1711.08488, 2017.

\bibitem{qi2017pointnet}
Charles~R Qi, Hao Su, Kaichun Mo, and Leonidas~J Guibas.
\newblock Pointnet: Deep learning on point sets for 3d classification and
  segmentation.
\newblock {\em Proc. Computer Vision and Pattern Recognition (CVPR), IEEE},
  1(2):4, 2017.

\bibitem{qi2016volumetric}
Charles~R Qi, Hao Su, Matthias Nie{\ss}ner, Angela Dai, Mengyuan Yan, and
  Leonidas~J Guibas.
\newblock Volumetric and multi-view cnns for object classification on 3d data.
\newblock In {\em Proceedings of the IEEE conference on computer vision and
  pattern recognition}, pages 5648--5656, 2016.

\bibitem{qi2017pointnet++}
Charles~Ruizhongtai Qi, Li Yi, Hao Su, and Leonidas~J Guibas.
\newblock Pointnet++: Deep hierarchical feature learning on point sets in a
  metric space.
\newblock In {\em Advances in Neural Information Processing Systems}, pages
  5099--5108, 2017.

\bibitem{redmon2016you}
Joseph Redmon, Santosh Divvala, Ross Girshick, and Ali Farhadi.
\newblock You only look once: Unified, real-time object detection.
\newblock In {\em Proceedings of the IEEE conference on computer vision and
  pattern recognition}, pages 779--788, 2016.

\bibitem{redmon2017yolo9000}
Joseph Redmon and Ali Farhadi.
\newblock Yolo9000: better, faster, stronger.
\newblock In {\em Proceedings of the IEEE conference on computer vision and
  pattern recognition}, pages 7263--7271, 2017.

\bibitem{redmon2018yolov3}
Joseph Redmon and Ali Farhadi.
\newblock Yolov3: An incremental improvement.
\newblock {\em CoRR}, abs/1804.02767, 2018.

\bibitem{ren2015faster}
Shaoqing Ren, Kaiming He, Ross Girshick, and Jian Sun.
\newblock Faster r-cnn: Towards real-time object detection with region proposal
  networks.
\newblock In {\em Advances in neural information processing systems}, pages
  91--99, 2015.

\bibitem{riegler2017octnet}
Gernot Riegler, Ali~Osman Ulusoy, and Andreas Geiger.
\newblock Octnet: Learning deep 3d representations at high resolutions.
\newblock In {\em Proceedings of the IEEE Conference on Computer Vision and
  Pattern Recognition}, volume~3, 2017.

\bibitem{song2016deep}
Shuran Song and Jianxiong Xiao.
\newblock Deep sliding shapes for amodal 3d object detection in rgb-d images.
\newblock In {\em Proceedings of the IEEE Conference on Computer Vision and
  Pattern Recognition}, pages 808--816, 2016.

\bibitem{song2017semantic}
Shuran Song, Fisher Yu, Andy Zeng, Angel~X Chang, Manolis Savva, and Thomas
  Funkhouser.
\newblock Semantic scene completion from a single depth image.
\newblock In {\em Computer Vision and Pattern Recognition (CVPR), 2017 IEEE
  Conference on}, pages 190--198. IEEE, 2017.

\bibitem{su2015multi}
Hang Su, Subhransu Maji, Evangelos Kalogerakis, and Erik Learned-Miller.
\newblock Multi-view convolutional neural networks for 3d shape recognition.
\newblock In {\em Proceedings of the IEEE international conference on computer
  vision}, pages 945--953, 2015.

\bibitem{su20153d}
Hao Su, Fan Wang, Eric Yi, and Leonidas~J Guibas.
\newblock 3d-assisted feature synthesis for novel views of an object.
\newblock In {\em Proceedings of the IEEE International Conference on Computer
  Vision}, pages 2677--2685, 2015.

\bibitem{xu2018multi}
Bin Xu and Zhenzhong Chen.
\newblock Multi-level fusion based 3d object detection from monocular images.
\newblock In {\em Proceedings of the IEEE Conference on Computer Vision and
  Pattern Recognition}, pages 2345--2353, 2018.

\bibitem{xu2017pointfusion}
Danfei Xu, Dragomir Anguelov, and Ashesh Jain.
\newblock Pointfusion: Deep sensor fusion for 3d bounding box estimation.
\newblock {\em CoRR}, abs/1711.10871, 2017.

\bibitem{yan2018second}
Yan Yan, Yuxing Mao, and Bo Li.
\newblock Second: Sparsely embedded convolutional detection.
\newblock {\em Sensors}, 18(10):3337, 2018.

\bibitem{yang2018hdnet}
Bin Yang, Ming Liang, and Raquel Urtasun.
\newblock Hdnet: Exploiting hd maps for 3d object detection.
\newblock In {\em Conference on Robot Learning}, pages 146--155, 2018.

\bibitem{yang2018pixor}
Bin Yang, Wenjie Luo, and Raquel Urtasun.
\newblock Pixor: Real-time 3d object detection from point clouds.
\newblock In {\em Proceedings of the IEEE Conference on Computer Vision and
  Pattern Recognition}, pages 7652--7660, 2018.

\bibitem{zhou2017voxelnet}
Yin Zhou and Oncel Tuzel.
\newblock Voxelnet: End-to-end learning for point cloud based 3d object
  detection.
\newblock {\em CoRR}, abs/1711.06396, 2017.

\bibitem{zhu2014single}
Menglong Zhu, Konstantinos~G Derpanis, Yinfei Yang, Samarth Brahmbhatt, Mabel
  Zhang, Cody Phillips, Matthieu Lecce, and Kostas Daniilidis.
\newblock Single image 3d object detection and pose estimation for grasping.
\newblock In {\em Robotics and Automation (ICRA), 2014 IEEE International
  Conference on}, pages 3936--3943. IEEE, 2014.

\end{thebibliography}
}

\end{document}